%% file: example_paper.tex
\documentclass{article}

\usepackage{microtype}
\usepackage{graphicx}
\usepackage{subfigure}
\usepackage{booktabs} 
\usepackage{enumitem}

\usepackage{hyperref}

\usepackage[accepted]{icml2025}

\usepackage{amsmath}
\usepackage{amssymb} 
\usepackage{mathtools}
\usepackage{amsthm}

\usepackage[capitalize,noabbrev]{cleveref}

\usepackage{subfigure}
\usepackage{caption}

\input{cmd} 
\theoremstyle{plain}
\newtheorem{theorem}{Theorem}[section]

\theoremstyle{definition}
\newtheorem{definition}[theorem]{Definition}

\theoremstyle{remark}

\usepackage[textsize=tiny]{todonotes}

\icmltitlerunning{TINED: GNNs-to-MLPs by Teacher Injection and Dirichlet Energy Distillation}

\begin{document}

\twocolumn[
\icmltitle{TINED: GNNs-to-MLPs by Teacher Injection and Dirichlet Energy Distillation}




\begin{icmlauthorlist}
\icmlauthor{Ziang Zhou}{polyu}
\icmlauthor{Zhihao Ding}{polyu}
\icmlauthor{Jieming Shi$^*$}{polyu}
\icmlauthor{Qing Li}{polyu}
\icmlauthor{Shiqi Shen}{tencent}
\end{icmlauthorlist}

\icmlaffiliation{polyu}{Department of Computing, The Hong Kong Polytechnic University, Hong Kong SAR, China}
\icmlaffiliation{tencent}{Wechat, Tencent, Beijing, China}

\icmlcorrespondingauthor{$^*$Jieming Shi}{jieming.shi@polyu.edu.hk}




\icmlkeywords{xxx}

\vskip 0.3in
]



\printAffiliationsAndNotice{}  

\begin{abstract}
Graph Neural Networks (GNNs) are pivotal in graph-based learning, particularly excelling in node classification. However, their scalability is hindered by the need for multi-hop data during inference, limiting their application in latency-sensitive scenarios. Recent efforts to distill GNNs into multi-layer perceptrons (MLPs) for faster inference often underutilize the layer-level insights of GNNs. In this paper, we present \algo, a novel approach that distills GNNs to MLPs on a layer-by-layer basis using Teacher Injection and Dirichlet Energy Distillation techniques.
We focus on two key operations in GNN layers: feature transformation (FT) and graph propagation (GP). We recognize that FT is computationally equivalent to a fully-connected (FC) layer in MLPs. Thus, we propose directly transferring teacher parameters from an FT in a GNN to an FC layer in the student MLP, enhanced by fine-tuning. In \algo, the FC layers in an MLP replicate the sequence of FTs and GPs in the GNN. We also establish a theoretical bound for GP approximation.
Furthermore, we note that FT and GP operations in GNN layers often exhibit opposing smoothing effects: GP is aggressive, while FT is conservative. Using Dirichlet energy, we develop a DE ratio to measure these effects and propose Dirichlet Energy Distillation to convey these characteristics from GNN layers to MLP layers. Extensive experiments show that \algo outperforms GNNs and leading distillation methods across various settings and seven datasets. Source code are available at \url{https://github.com/scottjiao/TINED_ICML25/}.

\end{abstract}

\input{contents/main_contents}

\end{document}

%% file: contents/main_contents.tex
\input{contents/introV2}

\input{contents/related}

\input{contents/preliminary}

\input{contents/method}

\input{contents/exp.tex}

\input{contents/conclusion}

\section*{Acknowledgements}

This work is supported by grants from the Research Grants Council of Hong Kong Special Administrative Region, China (No. PolyU 25201221, No. PolyU 15205224), and NSFC No. 62202404.
This project has been supported by the Hong Kong Research Grants Council under General Research Fund (project no. 15200023) as well as Research Impact Fund (project no. R1015-23).
This work is supported by Otto Poon Charitable Foundation Smart Cities Research Institute (SCRI) P0051036-P0050643, and grant P0048511 from Tencent Technology Co., Ltd.

\section*{Impact Statement}
This paper focuses on GNNs-to-MLPs distillation, which is a fundamental problem. There are many potential societal consequences of our work, none of which we feel must be specifically highlighted here.

\bibliography{example_paper}

\clearpage

\clearpage
\input{contents/appendix.tex}

%% file: contents/introV2.tex
\section{Introduction}
Graph neural networks (GNNs)  have  delivered impressive outcomes in important applications~\cite{graphsage, gcn, gat}.
The power of GNNs is underpinned by the message passing framework that assimilates and refines node representations by considering their (multi-hop) neighborhood over graphs~\cite{gnn_survey_1, gnn_survey_2}. Nonetheless, the message passing is computationally demanding due to numerous nodes involved, posing significant challenges to deploy GNNs in  latency-sensitive applications that require fast inference~\cite{agl, graphdec, jia2020redundancy}.

Recent studies attempt to combine the performance advantage of GNNs and the latency advantage of  multi-layer perceptrons (MLPs)~\cite{graph_mlp, glnn, nosmog, GA_MLP}.
Specifically, \glnn\cite{glnn}  distills teacher GNNs into student MLPs via soft labels. Then the student is deployed for fast inference to approximate the performance of GNNs without expensive message passing on graphs.
\nosmog\cite{nosmog} further considers  graph structures, robustness, and node relations, while a new  graph representation space is learned in~\cite{vqgraph}.
Current research  often treats both teacher GNN and student MLP as monolithic model entities, focusing primarily on soft label outputs of the GNN for distillation, which overlooks the  intrinsic knowledge contained within fine-grained GNN layers. 

In this paper, we conduct an in-depth analysis to reveal important properties of the key operations in GNN layers and propose a novel method, \algo, distilling GNNs to MLPs layer-wise by {\underline T}eacher \underline{IN}jection and Dirichlet {\underline E}nergy {\underline D}istillation. 
The main ideas are explained below, and the detailed architecture of \algo is presented later.

\begin{figure*}[!t]
	\centering
		\includegraphics[width=0.8\textwidth]{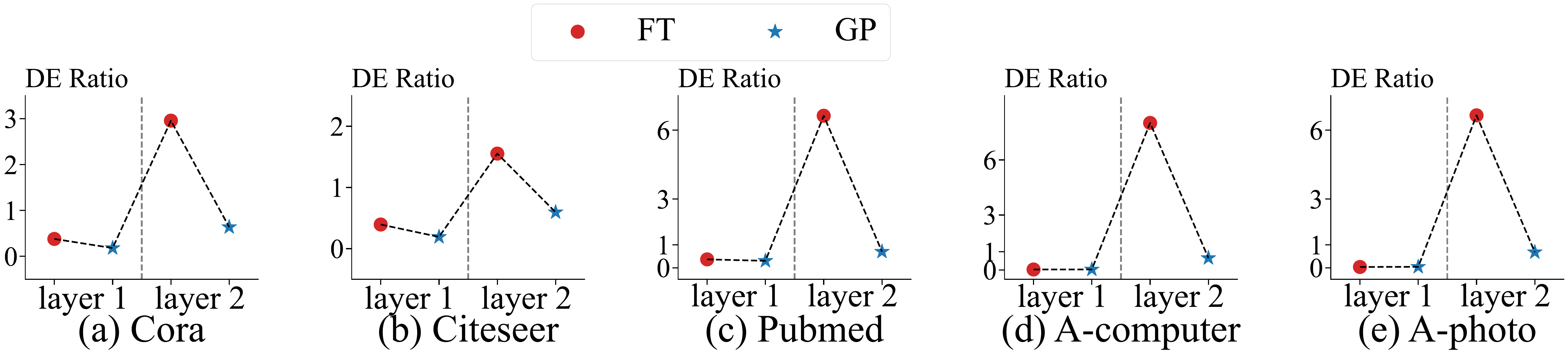}
   \vspace{-1mm}
\caption{The DE ratios of FTs and GPs in the layers of \sage.}\label{fig:2_layerwise_DE_ratio_sage}
\end{figure*}

\noindent\textbf{Main Idea of Teacher Injection.}
A typical GNN layer includes two key operations: feature transformation (FT) and graph propagation (GP) \cite{gcn,gat,graphsage,liu2020towards,unifiedframework}. GP aggregates neighbor representations of a node, while FT transforms node representations with learnable parameters. The valuable knowledge of a teacher GNN is preserved in the well-trained parameters of its FT and GP operations. Existing studies do not directly utilize these parameters~\cite{glnn}.
In Section \ref{sec:tin}, we recognize that \textit{FT operations in GNNs share the same formulation as fully-connected (FC) layers in MLPs}, both transforming representations using learnable transformation matrices and activations. Thus, we propose to inject the parameters of an FT in a teacher GNN layer into an FC layer in student MLP, followed by fine-tuning for distillation. The GP operation in the GNN layer is emulated by another FC layer in the student. This allows \algo to directly transplant teacher knowledge into the student. Moreover, in \algo, the FC layers in the MLP mirror the order of their corresponding FTs and GPs in the GNN teacher, preserving layer-specific knowledge as much as possible.
Theoretically, we prove an approximation bound between a GP in GNN and its corresponding FC layer in MLP, which depends on the eigenvalue of graph Laplacian matrix.

\noindent\textbf{Main Idea of Dirichlet Energy Distillation.} We further investigate the smoothing properties of FT and GP operations within GNN layers, since it is well-recognized that appropriate smoothing is crucial for  GNNs~\citep{measuringSmoothingAndRelieving}.
Importantly, we observe that \textit{the FT and GP operations in a GNN layer often exert opposing smoothing effects: GP aggressively smooths node embeddings, while FT is more restrained and can even diversify embeddings.}
Using Dirichlet energy~\cite{oversmoothingDEmeasure}, we propose a \textit{DE ratio} measure to quantify whether an operation is conservative (large DE ratio) or aggressive (small DE ratio) in smoothing. Figure \ref{fig:2_layerwise_DE_ratio_sage} shows the DE ratios of FTs and GPs in a trained 2-layer GNN teacher \sage on the experimental datasets. Within the same layer, the DE ratio of FT (red dot) is often larger than that of GP (blue star), indicating their opposing smoothing behaviors. Similar observations are made on other GNNs (see Appendix \ref{appendix:deratio}).
To distill these smoothing patterns from GNN  to MLP layers, we design Dirichlet Energy Distillation in Section \ref{sec:ded}.

We conduct extensive experiments on benchmark datasets under various settings. Results show that \algo achieves superior performance and fast  inference speed, compared with existing methods and various teacher GNNs.  For example, on the Citeseer data, \algo improves a GNN teacher by 3.94\%, MLPs by 15.93\%, \glnn by up to 3.21\%, and \nosmog by 1.54\%.  \algo is 94\texttimes\xspace faster than its GNN teacher for inference.
Our contributions are as follows:

\vspace{-\topsep-0.5pt}
 \begin{itemize}
 [topsep=-2pt,leftmargin=*]
  \setlength\itemsep{-2pt} 
  
\item We propose \algo, a novel method to effectively distill fine-grained layer-wise knowledge from teacher GNNs into student MLPs.

\item We develop a teacher injection technique to transplant the parameters of key operations from GNNs to MLPs. We provide a theoretical approximation analysis.

\item We observe distinct smoothing effects of FT and GP operations in GNN layers and introduce Dirichlet energy distillation to impart these effects to student MLPs.

\item Extensive experiments demonstrate that \algo achieves superior performance with various GNN teachers on widely-adopted benchmark datasets.

\end{itemize}

%% file: contents/related.tex
\section{Related Work}

GNNs leverage message-passing to aggregate neighborhood information for learning~\cite{gcn, gat, graphsage, GCNII, appnp}. GCN~\cite{gcn} introduces layer-wise propagation, GAT~\cite{gat} employs attention mechanisms, and APPNP~\cite{appnp} uses personalized PageRank. \citet{GCNII} mitigate over-smoothing with residual connections. Extensions include positional encoding~\cite{pgnn, peg, distance_encoding}, few-shot learning~\cite{ZhouSZH023, dong2025spacegnn}, and anomaly detection~\cite{dong2025smoothgnn, dong2023rayleigh, DZH2024}. Specialized GNNs, such as heterogeneous GNNs~\cite{ZhouSYZ023, hetgnn}, address specific graph types. Despite their effectiveness, GNN inference is computationally expensive.

\begin{figure*}[!t]
	\centering
		\includegraphics[width=0.84\textwidth]{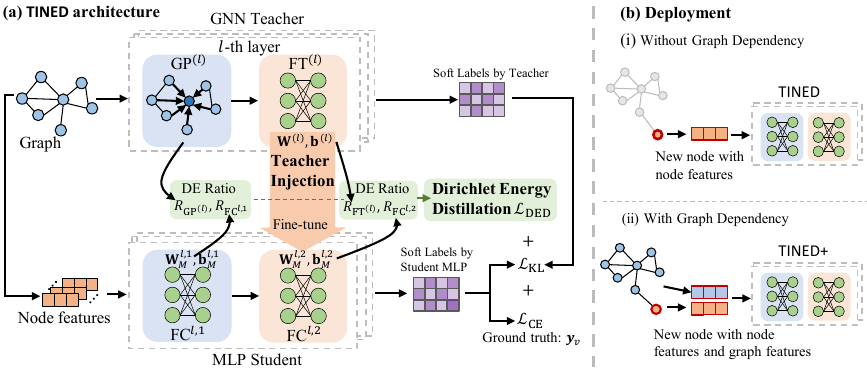}
   \vspace{-2mm}
 \caption{(a) \algo with Teacher Injection  and Dirichlet Energy Distillation; (b) Inference settings}
\label{fig:frame}   
\vspace{-3mm}
\end{figure*}

Knowledge Distillation has been applied in GNNs to accelerate inference while preserving effectiveness~\cite{lee2019graph, LSP, cpf_plp_ft, kg_on_graph_survey,freekd}. Previous studies have trained smaller student GNNs with fewer parameters than the large teacher GNNs, such as LSP~\cite{LSP}, FreeKD~\cite{freekd} and TinyGNN~\cite{tinygnn} which still rely on  time-intensive message passing.
GFKD~\cite{GFKD} performs distillation at graph-level instances, while we target nodes within a single graph.
Recent studies~\cite{nosmog,vqgraph, glnn, graph_mlp, cold_brew,ffg2m,krd} develop MLP-based student models without message passing.
\glnn~\cite{glnn} trains an MLP student using node features as input and soft labels from a GNN teacher as targets. 
\nosmog~\cite{nosmog} incorporates graph structure, adversarial feature augmentation, and node similarity relations into distillation. 
\vqgraph learns a structure-aware tokenizer to encode node substructures~\cite{vqgraph}, during which teacher is also trained.
As mentioned, the intrinsic knowledge in well-trained GNN layers is insufficiently utilized by existing studies.

There exist works in other orthogonal domains, also observing the benefits of distilling layer-wise teacher structures, such as language models~\cite{layerwise1}, natural language processing~\cite{layerwise1,layerwise2,kd_jiao} and computer vision~\cite{layerwise3,kd_hinton}. In this work, we fill the gap to develop layer-wise distillation method for GNNs. 
In addition, \citet{mlpinit} work on a different problem that uses pre-trained MLPs for GNN training acceleration.

%% file: contents/preliminary.tex
\section{Preliminaries}

\textbf{Notations.} 
A graph $\G = (\V, \E, \feat)$ consists of a node set $\V$ with $\numV$ nodes, i.e.,   $|\V|=\numV$, an edge set $\E$ of size  $|\E|=\numE$, and $\feat\in \mathbb{R}^{\numV \times \dimFeat}$ for $\dimFeat$-dimensional features of all nodes. 
Let $\adj$ be the adjacency matrix of $\G$, where $\adj_{u,v}=1$ if edge $(u,v)\in\E$, and 0 otherwise.
A node $v$ has a neighbor set $\nbr(v)=\{u|(u,v)\in \E\}$, and the degree of node $v$ is  
$|\nbr(v)|$. 
Degree matrix $\DM$ is a diagonal matrix with $\DM_{v,v}=|\nbr(v)|$.
In node classification task, 
the prediction targets are $\gtV\in \mathbb{R}^{\numV\times\numC}$, where $\numC$ is the number of classes, and row $\gt_v$ is a $\numC$-dim one-hot vector indicating the ground-truth class of node $v$.
A model predicts the class probabilities of node $v$.
In $\G$, we use superscript $^L$ to mark the labeled nodes (i.e., $\V^L$, $\feat^L$, and $\gtV^L$), and superscript $^U$ to mark the unlabeled nodes (i.e., $\V^U$, $\feat^U$, and $\gtV^U$).

\noindent\textbf{GNNs.}
Let $\HM^\la\in\mathbb{R}^{\numV\times d_l}$ be the output node embedding matrix of the $l$-th GNN layer, with each row $\h_v^\la$ being the representation of node $v$ in $\V$.
Most GNNs fit under the message-passing framework with feature transformation (FT) and graph propagation (GP) operations~\cite{unifiedframework}.
The $l$-th layer usually comprises operations $\ft^\la$  and $\gp^\la$ to get $\HM^\la$ in Eq. \eqref{eq:gnn_general}. 
$\gp^\la$ aggregates the $(l-1)$-th layer representations $\HM^\lminus$ over $\G$, and $\ft^\la$ transforms representations to get $\HM^\la$.  Different GNNs may vary in how they composite  $\ft^\la$ and $\gp^\la$. 
\begin{equation}\label{eq:gnn_general}
\small
\textstyle
\HM^\la= \ft^\la \left(\gp^\la \left(\HM^\lminus,\G\right)\right)
\end{equation}
\textbf{MLPs.}
An MLP is  composed of multiple fully-connected  (FC) layers. The $l$-th FC layer converts an embedding $\h^\lminus$ to $\h^\la$ via a transformation matrix $\WM_M^\la$ in Eq. \eqref{eq:mlplayer}.
\begin{equation}\label{eq:mlplayer}
\small
\h^{(l)}=\sigma(\h^{(l-1)}\WM_M^{(l)}+\mathbf{b}^\la),
\end{equation}
where $\sigma$ is an activation and $\mathbf{b}^\la$ is bias.

\noindent\textbf{GNNs-to-MLPs Distillation.}
Given a pre-trained GNN teacher, the goal is to train a cost-effective student MLP to predict $\predS_v$ for node $v$, utilizing ground-truth labels $\gt_v$ of labeled nodes $v \in \V^L$ and soft labels $\softT_v$ produced by teacher GNN for all $v \in \V$.
The training objective is formulated below \cite{glnn}. 
Note that the student is trained by both ground truth and soft labels from the teacher.
\begin{equation} 
\label{eq:basic_kd}
\small
\textstyle
    \mathcal{L} = \sum_{v \in \V^L} \mathcal{L}_{CE}( \predS_v,\gt_v) + \lambda \sum_{v \in \V } \mathcal{L}_{KL}(\predS_v, \softT_v),
\end{equation}
where $\mathcal{L}_{CE}$ is the cross-entropy loss by comparing the student MLP predictions $\predS_v$ with the ground truth $\gt_v$, and $\mathcal{L}_{KL}$ is the KL-divergence loss  between the student MLP predictions $\predS_v$ and the soft labels $\softT_v$ from teacher, and  weight $\lambda$ balances the two losses.

%% file: contents/method.tex
\section{The {TINED} Method}\label{sec:method}

In Eq. \eqref{eq:basic_kd}, existing studies primarily use GNN soft labels and ground truth labels for distillation. Our approach \algo  expands this to include model parameters and layer-specific properties of teacher GNNs for distilling knowledge into MLPs, as illustrated in
Figure \ref{fig:frame}.
As shown in Figure \ref{fig:frame}(a), for the Teacher Injection technique,   we identify that the  $\ft^\la$ operation of the  $l$-th GNN layer  essentially conducts the same computation as an FC layer in an MLP, and thus, we inject the parameters of $\ft^\la$ into an FC layer $\fc^\latwo$ of the student MLP. The injected parameters are fine-tuned during the distillation to ensure controlled adaption.
To emulate the $\gp^\la$ operation in the $l$-th GNN layer, we employ another FC layer $\fc^\laone$ in the student MLP. We establish  a  theoretical approximation bound between $\gp^\la$ and $\fc^\laone$.
Then  as in Figure \ref{fig:frame}(a), we use Dirichlet energy to develop DE ratios $\der_{\gp^\la}$ and $\der_{\ft^\la}$ to quantify the smoothing effects of $\gp^\la$ and $\ft^\la$ in the  $l$-th GNN layer, and design Dirichlet Energy Distillation with a $\mathcal{L}_{DED}$ loss to ensure the DE ratios $\der_{\fc^\latwo}$ and $\der_{\fc^\laone}$ of the student MLP's FC layers $\fc^\laone$ and $\fc^\latwo$ preserve the smoothing effects of $\gp^\la$ and $\ft^\la$ from the teacher GNN.

There are two typical  inference deployment settings  as depicted in Figure \ref{fig:frame}(b). 
\algo can operate without graph dependency for inference~\cite{glnn}, which is useful when the graph structure is unavailable or when new unseen nodes lack graph connections; \algo  can also work with graph structures when allowed~\cite{nosmog}.  
Following~\cite{glnn,nosmog}, we use GraphSAGE~\citep{graphsage} with GCN aggregation  as the teacher GNN to explain \algo. Our experiments include results over different teacher GNNs.
In Appendix \ref{app:otherTeacherGNNs}, we explain how to use \algo for various GNNs. 

\subsection{Teacher Injection}\label{sec:tin}

The idea is to inject the parameters of certain operations of GNNs into MLPs, thereby directly transferring the teacher's knowledge to the student.
The GNN operations to be injected should be {compatible} with MLP in Eq. \eqref{eq:mlplayer}, i.e., with the same formulation.
As shown in Eq. \eqref{eq:gnn_general}, a GNN layer consists of GP and FT operations. In what follows, we show that the FT operations of GNNs can be injected into student MLPs.
We use \sage as  an example, while the analysis on other GNNs as teacher is in Appendix  \ref{app:otherTeacherGNNs}.
The $l$-th layer of \sage with  $\gp^\la$ and $\ft^\la$ operations is
\begin{equation}\label{eq:sage}
\small
\begin{aligned}
\gp^\la\text{: } &\tilde{\h}_{v}^\la = \concat\left(\h_v^\lminus,\agg^{\la}\left(\{\h_u^\lminus, \forall u \in \nbr(v)\}\right)\right),\\
    \ft^\la\text{: } &\h_v^\la = \sigma \left(\tilde{\h}_{v}^\la\cdot \WM^\la +\mathbf{b}^\la \right).
\end{aligned}
\end{equation}
In   Eq. \eqref{eq:sage}, $\gp^\la$ operation has an aggregator $\agg^\la$ to combine the $\lminus$-th representations $\h_u^\lminus$ of $v$'s neighbors and then concatenates (\concat)  them with $\h_v^\lminus$ to get $\tilde{\h}_{v}^\la$. In $\ft^\la$ operation, a learnable transformation matrix $\WM^\la$ is applied to $\tilde{\h}_{v}^\la$, followed by an activation function to yield the $l$-th layer representation $\h_v^\la$ of $v$.

Observe that (i) $\ft^\la$ operates independently of the graph, whereas $\gp^\la$ requires the graph; (ii) $\ft^\la$ in Eq. \eqref{eq:sage} has the same formulation as an FC layer of an MLP in Eq. \eqref{eq:mlplayer}.
Therefore, to approximate the $l$-th GNN layer, we employ two FC layers $\fc^\laone$ and $\fc^\latwo$ in the student MLP. 
In Eq. \eqref{eq:studentlayers}, the $\fc^\laone$ layer of the student MLP is to approximate $\gp^\la$ in Eq. \eqref{eq:sage}, while the $\fc^\latwo$ layer approximates $\ft^\la$ in Eq. \eqref{eq:sage}, with teacher injection to set  $\WM_M^\latwo=\WM^\la$ and $\mathbf{b}_M^\latwo=\mathbf{b}^\la$, transferring the parameters of $\ft^\la$ to $\fc^\latwo$.
$\s_v^\la$ is the embedding of node $v$ generated by the student.
\begin{equation}\label{eq:studentlayers}
\small
\begin{aligned}
\fc^{\laone} {\text{ for }}\gp^{\la}: \hat{\s}_{v}^\la=\sigma\left(\s^\prime_{v}\WM_M^{\laone}+\mathbf{b}_M^{\laone}\right), \\
\fc^{\latwo} {\text{ for }}\ft^{\la}:\s_{v}^\la=\sigma\left(\hat{\s}_{v}\WM_M^{\latwo}+\mathbf{b}_M^{\latwo}\right),
\end{aligned}
\end{equation}
where $\WM_M^\latwo=\WM^\la$  and  $\mathbf{b}_M^\latwo=\mathbf{b}^\la$, and   $\WM^\la$ and $\mathbf{b}^\la$ are from Eq. \eqref{eq:sage}.

Different from existing methods, \algo distills knowledge from  GNNs to MLPs on a per-layer basis. Directly injecting GNN parameters into the MLP is anticipated to improve the effectiveness. 
For layer-wise GNNs with $T$ layers, the resulting MLP will have $2T$ FC layers. Typically, $T$ is small, e.g., 2 or 3, and the efficiency impact on inference between MLPs with 2 and 4 layers is minimal (1.45ms vs. 1.63ms, as reported in~\cite{glnn}, compared to 153ms by \sage). 
Furthermore, our experiments show that \algo can achieve a favorable tradeoff between effectiveness and efficiency for inference. 
For decoupled GNNs, e.g., APPNP \cite{appnp}, with FTs and GPs decoupled, the teacher injection is also applicable with a small number of FC layers (see Appendix \ref{app:otherTeacherGNNs}).

In Theorem \ref{theoremErrorBound}, we provide a theoretical approximation bound between $\gp^\la$ in the $l$-th layer of teacher GNN and its corresponding $\fc^\laone$ in student MLP, serving as an attempt to establish the  relationship between GNNs and MLPs for knowledge distillation.
The proof is in Appendix~\ref{proof:errorbound}.

\begin{theorem}\label{theoremErrorBound}
    For a sparse matrix $\LM\in \mathbb{R}^{\numV\times \numV}$
    and a feature matrix $\mathbf{H}\in \mathbb{R}^{\numV\times d}$ with ${rank}(\mathbf{H})=d$, there exists a transformation matrix $\mathbf{W}^*$ to approximate $\LM\HM$ by $\HM\mathbf{W}^*$ with relative error 
\begin{equation*}
    \small
    \frac{||\mathbf{L}\mathbf{H}-\mathbf{H}\mathbf{W}^*||_F}{||\mathbf{H}||_F}   \leq \lambda_{\max}(\mathbf{L}),
\end{equation*} where $||\cdot||_F$ is the Frobenius norm and $ \lambda_{\max}(\mathbf{L})$ is the largest eigenvalue of laplacian $\mathbf{L}$.
\end{theorem}

Specifically, Laplacian matrix $\LM$ represents graph topology used in $\gp^\la$, $\LM\HM$ represents the output of $\gp^\la$, while $\HM\WM^*$ represents the output of $\fc^\laone$ and $\HM\WM^*$ approximates $\LM\HM$. 
The relative error is upper-bounded by the largest eigenvalue $\lambda_{\max}(\mathbf{L})$ of $\LM$, regardless of graph size.

The parameters of  FC layers  $\fc^\latwo$ in Eq. \eqref{eq:studentlayers} are sourced from the $\ft^\la$ operations of  teacher GNN. To ensure these FC layers contribute to the student MLP's training in a regulated manner, we introduce a gradient modifier $\eta$ to fine-tune the injected parameters when updating the gradients,
\begin{equation}\label{eq:fine_tuning}
\small
\begin{aligned}
    \hat{\nabla} \WM_M^{\latwo}=&\eta \nabla \WM_M^{\latwo}\\
    \hat{\nabla} \mathbf{b}_M^{\latwo}=&\eta \nabla \mathbf{b}_M^{\latwo},
\end{aligned}
\end{equation}
where $\eta$ is a hyperparameter to control fine-tuning, while $\nabla$ represents gradient. 

In other words, during training, the update of the parameters in $\fc^\latwo$ based on gradient optimization after backward propagation is adjusted by $\eta$ parameter.

\subsection{Dirichlet Energy Distillation}\label{sec:ded}
It is widely recognized that an appropriate degree of smoothing is crucial for the efficacy of GNNs. Here we make an interesting observation  that, within the $l$-th GNN layers, $\gp^\la$ operations tend to aggressively smooth node embeddings, while $\ft^\la$ operations often apply conservative smoothing or even diversify embeddings. These distinct layer-level smoothing behaviors in teacher GNNs should be captured in the student MLPs.

To achieve this, we introduce Dirichlet Energy Distillation.
Dirichlet energy~\cite{oversmoothingDEmeasure} is a measure commonly used to quantify the degree of smoothing in embeddings by evaluating node pair distances.
Definition \ref{def:DE} defines the Dirichlet energy $\DE(\HM)$ of a node embedding matrix $\HM$.

\begin{definition}\label{def:DE}
    Given a node embedding matrix $\HM\in\mathbb{R}^{\numV\times d}$, learned from either the GNN teacher or the MLP student at a certain layer, the Dirichlet energy of $\HM$ is 
    \begin{equation}
      \small  \DE(\HM) = {\frac{1}{n}\cdot\text{tr}(\HM^\top\LM\HM)},
    \end{equation}
    where $\text{tr}(\cdot)$ is the trace of a matrix, and $\LM=\mathbf{D}-\mathbf{A}$ is the Laplacian matrix of graph $\G$.
\end{definition}

A lower Dirichlet energy value $\DE(\HM)$ suggests that the embeddings in $\HM$ are smooth, whereas a higher value indicates diversity among the embeddings. For an operation $\op$ that processes $\HM$ to output $\op(\HM)$, if  $\DE(\op(\HM)) < \DE(\HM)$, then the operation $\op$ is smoothing the embeddings; on the other hand, if $\DE(\op(\HM)) > \DE(\HM)$, $\op$ diversifies them. 

We define Dirichlet energy ratio $\der_{\op}$ (DE ratio) as follows.
\begin{definition}[DE ratio]
    The DE ratio $\der_{\op}$ of an operation $\op$ is the Dirichlet energy of its output $\DE(\op(\HM))$ over the Dirichlet energy of its input $\DE(\HM)$,  
$\der_{\op}=\frac{\DE(\op(\HM))}{\DE(\HM)}$.
\end{definition}

On a trained teacher \sage with 2 layers, in each layer, we calculate the DE ratios $\der_{\ft^\la}$ and $\der_{\gp^\la}$ of $\ft^\la$ and $\gp^\la$ operations respectively. We repeat  10 times and report the average DE ratios in Figure \ref{fig:2_layerwise_DE_ratio_sage}, over the benchmark datasets used in experiments. 

While the overall Dirichlet energy of embeddings is decreasing in \sage, in a specific $l$-th layer shown in Figure \ref{fig:2_layerwise_DE_ratio_sage}, we make the following two consistent observations about {DE ratio}. (i) Within the same layer for $l=1,2$, the DE ratio $\der_{\ft^\la}$ of $\ft^\la$ consistently exceeds the DE ratio $\der_{\gp^\la}$ of $\gp^\la$, demonstrating that $\gp^\la$ operation actively smooths embeddings, whereas $\ft^\la$ operation is relatively conservative for smoothing. (ii) At $l=2$, DE ratio $\der_{\ft^\la}$ even surpasses 1, indicating that in this layer, $\ft^\la$ acts to diversify embeddings rather than smoothing them. These trends are consistently observed across layer-wise GNNs, including GCN and GAT in Appendix \ref{appendix:deratio}.

Figure \ref{fig:2_layerwise_DE_ratio_sage} reveals that  $\ft$s and $\gp$s can have opposing effects on smoothing. 
Recall that we associate $\ft^\la$ and $\gp^\la$ of the $l$-th GNN layer with  $\fc^\latwo$ and $\fc^\laone$ layers in the student MLP in Eq. \eqref{eq:studentlayers}. The proposed Dirichlet Energy Distillation technique aims to encapsulate the distinct smoothing behaviors of $\ft^\la$ and $\gp^\la$ into $\fc^\latwo$ and $\fc^\laone$ respectively, thereby transferring teacher GNN's knowledge of smoothing effects to student MLP.

Specifically, the loss of the $l$-th GNN layer for Dirichlet Energy Distillation, $\mathcal{L}_{DED}^\la$, is the sum of the squares of the difference between DE ratios $\der_{\gp^\la}$ of $\gp^\la$ and  $\der_{\fc^\laone}$ of the $\fc^\laone$  layer, and the difference between $\der_{\ft^\la}$ of $\ft^\la$ and $\der_{\fc^\latwo}$ of the $\fc^\latwo$ layer.
The total DED loss $\mathcal{L}_{DED}$ is the sum of $\mathcal{L}_{DED}^\la$ for all GNN layers $l=1,..,\numL$.
\begin{equation}
\small
\begin{aligned}
&\mathcal{L}_{DED}^\la=(\der_{\gp^\la}-\der_{\fc^\laone})^2+(\der_{\ft^\la}-\der_{\fc^\latwo})^2\\
&\mathcal{L}_{DED}=\sum_{l=1}^\numL \mathcal{L}_{DED}^\la.
\end{aligned}
\end{equation}

The final objective function $\mathcal{L}$ of our method \algo is the weighted combination of ground truth cross-entropy
loss $\mathcal{L}_{CE}$, soft label distillation loss $\mathcal{L}_{KL}$, and Dirichlet Energy Distillation loss  $\mathcal{L}_{DED}$:
\begin{equation}\label{eq:total_loss}
\small
    \mathcal{L} = \sum_{v \in \V^L} \mathcal{L}_{CE}( \predS_v,\gt_v) + \lambda \sum_{v \in \V } \mathcal{L}_{KL}(\predS_v, \softT_v) + \beta\sum_{l=1}^\numL \mathcal{L}_{DED}^\la,
\end{equation}
where $\lambda$ and $\beta$ are weights for balancing the loss functions.

%% file: contents/exp.tex
\section{Experiments}
\label{sec:experiments}

\subsection{Experiment Settings}\label{sec:expSetting}
\textbf{Datasets.} 
We use 7 widely used public benchmark datasets, including  \cora{}, \citseer{}, \pubmed{}, \acomputer{}, \aphotos{}~\cite{glnn, cpf_plp_ft}, and \ogba{} and \ogbp~\cite{ogb}  that are two large OGB datasets, to evaluate our method and baselines. Table \ref{tab:dataset} in Appendix \ref{app:data} provides the data statistics and splits.

\begin{table*}[!t]
    \center
    \small
    \caption{Under the {transductive}(\transductive) setting, the accuracy results of the teacher and all methods without or with graph dependency for inference on  $\V^U$ (higher is better). The best in each category is in bold. OOM means out-of-memory.
    }
    \vspace{-0mm}
    \renewcommand{\arraystretch}{1}

    \begin{adjustbox}{width=1\textwidth,center}
    \setlength{\tabcolsep}{4pt}
    \begin{tabular}{l|l|llllll|lll}
    \toprule
 & Teacher& \multicolumn{6}{c|}{Without Graph Dependency}& \multicolumn{3}{c}{With Graph Dependency}\\
    \midrule
    \multicolumn{1}{l|}{Datasets} & \multicolumn{1}{l|}{SAGE} & MLP & {FFG2M} & {KRD} & GLNN  &GLNN$^*$& \algo & NOSMOG& NOSMOG$^*$& \algog\\ \midrule
    \cora & 80.64$\pm$1.57 & 59.18$\pm$1.60 &82.38$\pm$1.41 &82.27$\pm$1.31 & 80.26$\pm$1.66  & 81.31$\pm$1.62&\textbf{82.63$\pm$1.57}& 83.04$\pm$1.26& 82.27$\pm$1.75& \textbf{83.70$\pm$1.02}\\
    \citseer & 70.49$\pm$1.53 & 58.50$\pm$1.86 &72.85$\pm$1.59 &72.84$\pm$1.70 & 71.22$\pm$1.50  & 72.38$\pm$1.40&\textbf{74.43$\pm$1.53}& 73.78$\pm$1.54& 73.85$\pm$2.27& \textbf{75.39$\pm$1.59}\\
    \pubmed & 75.56$\pm$2.06 & 68.39$\pm$3.09 &76.56$\pm$3.41 &77.01$\pm$3.11 & 75.59$\pm$2.46  & 76.95$\pm$2.72&\textbf{77.09$\pm$2.14}& 77.34$\pm$2.36& 76.79$\pm$2.65& \textbf{77.75$\pm$3.14}\\
    \acomputer & 82.82$\pm$1.37 & 67.62$\pm$2.21 &83.67$\pm$1.04 &82.87$\pm$0.87 & 82.71$\pm$1.18  & 83.64$\pm$1.13&\textbf{85.18$\pm$1.12}& 84.04$\pm$1.01& 84.33$\pm$1.14& \textbf{84.82$\pm$1.58}\\
    \aphotos & 90.85$\pm$0.87 & 77.29$\pm$1.79 &93.18$\pm$0.87 &92.82$\pm$0.74 & 91.95$\pm$1.04  & 92.99$\pm$0.63&\textbf{93.97$\pm$0.53}& 93.36$\pm$0.69& 93.57$\pm$0.48& \textbf{94.05$\pm$0.39}\\
    \ogba & 70.73$\pm$0.35 & 55.67$\pm$0.24 &58.51$\pm$0.35 &59.26$\pm$0.51 & 63.75$\pm$0.48  & 63.78$\pm$0.69&\textbf{64.44$\pm$0.72}& \textbf{71.65$\pm$0.29}& 71.17$\pm$0.60& 71.52$\pm$0.34  \\
    \ogbp & 77.17$\pm$0.32 & 60.02$\pm$0.10 &OOM &OOM & 63.71$\pm$0.31  & 65.56$\pm$0.26&\textbf{69.48$\pm$0.25}& 78.45$\pm$0.38& 78.47$\pm$0.28& \textbf{78.59$\pm$0.28}\\
    \bottomrule
    \end{tabular}
    \end{adjustbox}
    \label{tab:standard}
    \vspace{-0mm}
    \end{table*}

\noindent\textbf{Teacher Architectures.}
Following~\cite{glnn,nosmog},  for the main results, \sage~\cite{graphsage} with GCN aggregation is used as the teacher model. We also conduct experiments of different GNN teachers, e.g.,  GCN, GAT and APPNP in Section \ref{sec:different}. 
The details of teachers and hyperparameter settings are in  \ref{app:teacherhyper}.

\noindent\textbf{Methods.}
\algo does not require graph dependency for inference, similar to \glnn~\cite{glnn}.   When graph dependency is permissible, we enhance it to create \algog, akin to \nosmog~\cite{nosmog}, to incorporate graph structural information.
Baselines \glnn and \nosmog use their original layer and hidden dimension configurations as per their papers. Additionally, we configure them with the same layer and hidden dimension settings as ours, dubbed as \glnnstar and \nosmogstar, for evaluation. 
We further compare with KRD and FFG2M~\cite{krd,ffg2m}, as well as with MLPs and the  \sage teacher.

\noindent\textbf{Transductive and Inductive Settings.} Two settings are considered~\cite{glnn,nosmog}: transductive  (\transductive) setting, and production (\production) setting with  both inductive and transductive evaluations  (\inductive \& \transductive).

\noindent{\transductive}: A model is trained on $\gG$, $\mX$, and $\mY^L$, and soft labels $\softT_v$ of all nodes in $\V$ are used for knowledge distillation. Inference evaluation is conducted on
the nodes in {$\V^U$.}

\noindent  {\production (\inductive \& \transductive)}: we randomly select out 20\% of nodes from the unlabeled nodes  $\V^U$, dividing $\V^U$ into disjoint inductive (unobserved) subset and observed subset, $\V^U = \V^U_{obs} \sqcup \V^U_{ind}$.
Node features and labels are partitioned into disjoint sets, i.e. $\mX = \mX^L \sqcup \mX_{obs}^U \sqcup \mX_{ind}^U$, and $\mY = \mY^L \sqcup \mY_{obs}^U \sqcup \mY_{ind}^U$.
Let $\gG_{obs}$ be the graph induced from $\gG$, with edges connecting nodes in $\V^L\sqcup\V^U_{obs}$. 
A model is trained on  $\gG_{obs}$, $\mX^L$, $\mX^U_{obs}$, and $\mY^L$.
Soft labels  in 
subsets $\V^L\sqcup\V^U_{obs}$ are used for distillation. In this \production (\inductive\& \transductive) setting, inference is evaluated on  $\V^U_{ind}$ for \inductive and  $\V^U_{obs}$ for \transductive (which is different from the \transductive setting over  $\V^U$ explained above). The overall \production performance is the weighted sum of \inductive performance  on $\V^U_{ind}$ and \transductive on $\V^U_{obs}$, with weights proportaional to the size of $\V^U_{ind}$ and $\V^U_{obs}$.

\noindent\textbf{Evaluation.} 
Following \cite{nosmog,glnn}, we present the mean and standard deviation of performance results from 10 trials, each with a unique random seed. A method is evaluated by accuracy  with the best model chosen  on validation data and tested on test data.

\subsection{Performance in Transductive  Setting}\label{sec:exp:tran}
Table \ref{tab:standard} reports the results under the transductive setting with inference  on node set $\V^U$. 
Table \ref{tab:standard} can be directly comparable to those reported in literature~\cite{nosmog,glnn,ogb, cpf_plp_ft}. 
In Table \ref{tab:standard}, GNNs-to-MLPs methods are in two categories: with or without graph dependency for inference. Baselines \glnn and \nosmog use their original student MLP configurations for layers and hidden dimensions as recommended in their respective papers. \glnnstar and \nosmogstar (with an asterisk $^*$) adopt the same configurations as our model for a fair comparison.
\algo surpasses all baselines without graph dependency, often with a significant margin. 
For example, on the large OGB  \ogbp dataset, \algo achieves gains of  9.42\%, 5.77\%, and 3.92\% over MLP, \glnn, and \glnnstar, while baselines KRD and FFG2M are OOM. On \citseer, \algo achieves 74.43\% accuracy, improving by 15.93\%, 3.21\%,  2.05\%, 1.59\%, and 1.58\% over MLP, \glnn, \glnnstar, KRD, and FFG2M.
The performance of \algo validates the power of the proposed techniques to preserve layer-level knowledge into MLPs.
Similar to \glnn, \algo surpasses the teacher on the first five data, while being effective on the last two.
When graph dependency is allowed, \algog 
excels \nosmog and \nosmogstar on most datasets, except \ogba where the performance is comparable.
On \citseer, \algog achieves 75.39\% accuracy, 1.54\% higher than \nosmogstar.
Moreover, all methods with graph dependency, including \algog and \nosmog, perform better than those without graph, and \algog excels the teacher model on all datasets, indicating the extensibility of  \algog to consider graph structures for inference.

\begin{table*}[t]
    \centering
    \caption{
 In  \production (\inductive\& \transductive) setting, the accuracy results of the teacher and all methods without or with graph dependency for online inference are presented.    
 \inductive indicates the results on  $\V^U_{ind}$, \transductive indicates the results on $\V^U_{obs}$, and \production indicates the weighted sum of the performance of both \inductive and \transductive with weights proportional to the size of $\V^U_{ind}$ and $\V^U_{obs}$.
  The best result in each category is in bold. The teacher's performance surpassing the best result in at least one category is italicized. 
    }
    \vspace{-0mm}
    \setlength{\tabcolsep}{4pt}
    \renewcommand{\arraystretch}{1}

   \begin{adjustbox}{width=0.999\textwidth,center}
   
    \begin{tabular}{ll|l|llllll|lll}
    \toprule
 & & Teacher& \multicolumn{6}{c|}{Without Graph Dependency}& \multicolumn{3}{c}{With Graph Dependency}\\
    \midrule
    
    {Datasets} & Eval & SAGE & MLP & {FFG2M} & {KRD} & GLNN  &\glnnstar&
\algo & NOSMOG &NOSMOG*& \algog\\ \midrule

    \multirow{3}{*}{\cora} & \production & 79.53 & 59.18 & 78.60&  75.74& 77.82  & 78.14& \textbf{78.90}& \textbf{81.02}& 80.30& 80.77\\ 
            & \inductive & \textit{81.03$\pm$1.71} & 59.44$\pm$3.36 & 72.02$\pm$1.43& 70.26$\pm$1.94& 73.21$\pm$1.50  & 73.58$\pm$1.42& \textbf{74.38$\pm$1.28}& 81.36$\pm$1.53 & 80.98$\pm$2.39& \textbf{81.50$\pm$2.54}\\
            & \transductive & 79.16$\pm$1.60 & 59.12$\pm$1.49 & 80.01$\pm$1.41& 77.11$\pm$1.44&  78.97$\pm$1.56  & 79.65$\pm$1.45& \textbf{80.04$\pm$1.50}& \textbf{80.93$\pm$1.65} & 80.13$\pm$1.64& 80.59$\pm$1.45\\ \midrule
    \multirow{3}{*}{\citseer} & \production & 68.06 & 58.49 & 71.89&  71.38& 69.08  & 70.91& \textbf{72.29}& 70.60 & 71.33& \textbf{73.58}\\ 
            & \inductive & 69.14$\pm$2.99 & 59.31$\pm$4.56 & 69.75$\pm$3.16& 69.78$\pm$3.04&  68.48$\pm$2.38  & 71.10$\pm$1.50& \textbf{72.68$\pm$1.97}& 70.30$\pm$2.30 & 72.35$\pm$2.99& \textbf{74.20$\pm$1.67}\\
            & \transductive & 67.79$\pm$2.80 & 58.29$\pm$1.94 & 72.12$\pm$2.69& 71.77$\pm$2.81&  69.23$\pm$2.39  & 70.86$\pm$1.66& \textbf{72.20$\pm$1.66}& 70.67$\pm$2.25 & 71.07$\pm$1.85& \textbf{73.43$\pm$1.63}\\ \midrule
    \multirow{3}{*}{\pubmed} & \production & 74.77 & 68.39 & 73.98& \textbf{76.00}&  74.67  & 75.21& 75.79& 75.82 & 75.57& \textbf{75.90}\\ 
            & \inductive & 75.07$\pm$2.89 & 68.28$\pm$3.25 &  73.49$\pm$7.91&  75.17$\pm$3.11&  74.52$\pm$2.95  & 74.83$\pm$2.83& \textbf{75.64$\pm$3.02}& 75.87$\pm$3.32 & 75.49$\pm$2.96& \textbf{76.30$\pm$2.95}\\
            & \transductive & 74.70$\pm$2.33 & 68.42$\pm$3.06 &  74.10$\pm$7.78& \textbf{76.20$\pm$3.00}&  74.70$\pm$2.75  & 75.30$\pm$2.70& 75.83$\pm$2.81& 75.80$\pm$3.06 & 75.58$\pm$2.83& \textbf{75.80$\pm$2.88}\\ \midrule
    \multirow{3}{*}{\acomputer} & \production & 82.73 & 67.62 &  82.69& 81.17& 82.10  & 83.23& \textbf{84.46}& 83.85 & 85.02& \textbf{85.08}\\
            & \inductive & \textit{82.83$\pm$1.51} & 67.69$\pm$2.20 & 80.52$\pm$1.56& 79.15$\pm$1.82&  80.27$\pm$2.11  & 81.10$\pm$1.49& \textbf{82.83$\pm$1.45}& 84.36$\pm$1.57 & 85.23$\pm$1.51& \textbf{85.45$\pm$1.60}\\
            & \transductive & 82.70$\pm$1.34 & 67.60$\pm$2.23 &  83.23$\pm$1.36& 81.67$\pm$1.92& 82.56$\pm$1.80  & 83.77$\pm$1.36& \textbf{84.87$\pm$1.38}& 83.72$\pm$1.44 & 84.97$\pm$1.44& \textbf{84.98$\pm$1.32}\\ \midrule
    \multirow{3}{*}{\aphotos} & \production & 90.45 & 77.29 &  92.35& 91.84& 91.34  & 91.54& \textbf{93.38}& 92.47 & 92.00& \textbf{93.12}\\ 
            & \inductive & 90.56$\pm$1.47 & 77.44$\pm$1.50 & 90.70$\pm$0.76& 90.04$\pm$1.12&  89.50$\pm$1.12  & 89.35$\pm$0.89& \textbf{91.96$\pm$0.72}& 92.61$\pm$1.09 & 92.93$\pm$0.96& \textbf{93.27$\pm$0.85}\\
            & \transductive & 90.42$\pm$0.68 & 77.25$\pm$1.90 & 92.77$\pm$0.24& 92.29$\pm$0.63&  91.80$\pm$0.49  & 92.09$\pm$0.71& \textbf{93.74$\pm$0.51}& 92.44$\pm$0.51 & 91.77$\pm$0.69& \textbf{93.08$\pm$0.68}\\ \midrule
    \multirow{3}{*}{\ogba} & \production & \textit{70.69} & 55.35 & 59.60&  59.32& 63.50  & \textbf{64.17}&63.24& 70.90 & 70.95& \textbf{71.22}\\ 
            & \inductive & \textit{70.69$\pm$0.58} & 55.29$\pm$0.63 &  57.02$\pm$0.43& 57.32$\pm$0.31& 59.04$\pm$0.46  & 58.73$\pm$0.46&\textbf{59.79$\pm$0.46}& 70.09$\pm$0.55  & 70.12$\pm$0.39& \textbf{70.42$\pm$0.35}\\
            & \transductive & \textit{70.69$\pm$0.39} & 55.36$\pm$0.34 & 60.24$\pm$0.23& 59.82$\pm$0.27&  64.61$\pm$0.15  & \textbf{65.53$\pm$0.30}&64.10$\pm$0.38& 71.10$\pm$0.34 & 71.16$\pm$0.15& \textbf{71.43$\pm$0.19}\\ \midrule
    \multirow{3}{*}{\ogbp} & \production & \textit{76.93} & 60.02 & OOM&  OOM& 63.47  & 68.48&\textbf{69.35}& 77.33 & 78.25& \textbf{78.91}\\ 
            & \inductive & \textit{77.23$\pm$0.24} & 60.02$\pm$0.09 & OOM&OOM &  63.38$\pm$0.33  & 68.13$\pm$0.20& \textbf{68.68$\pm$0.27}& 77.02$\pm$0.19  & 78.52$\pm$0.22& \textbf{79.31$\pm$0.29}\\
            & \transductive & \textit{76.86$\pm$0.27} & 60.02$\pm$0.11 &OOM &OOM &  63.49$\pm$0.31  & 68.57$\pm$0.20& \textbf{69.52$\pm$0.27}& 77.41$\pm$0.21 & 78.18$\pm$0.23& \textbf{78.81$\pm$0.29}\\
    \bottomrule
    \end{tabular}
    \end{adjustbox}
    \label{tab:prod}
    \vspace{-0mm}
\end{table*}

\subsection{Performance in Production Setting with \inductive \& \transductive}
We then conduct experiments under the \production (\inductive \& \transductive) setting with results in Table \ref{tab:prod}.
Note that the \transductive results in Table \ref{tab:prod} are over  $\V^U_{obs}$, different from the   results in Table \ref{tab:standard} on  $\V^U$ in Section \ref{sec:exp:tran}.
In Table \ref{tab:prod}, \algo and \algog can outperform  the teacher model and the baseline methods in categories with or without graph dependency under almost all settings. 
For instance, without graph dependency for inference,  \algo is better  than baselines, except \ogba and \pubmed where close performance is achieved. 
As an example, on \acomputer under \inductive setting, \algo achieves 82.83\% accuracy, significantly improving \glnnstar by 1.73\%.
With graph dependency, \algog surpasses \nosmog methods on all datasets  except \production and \transductive setting on \cora, where the performance is comparable. 
For instance, on \citseer under \production, \algog achieves performance gain of 2.98\% and 2.25\% over \nosmog and \nosmogstar respectively.
Moreover, on \ogba and \ogbp with a significant distribution shift between training and test data~\cite{glnn}, \algog outperforms the teacher and \nosmog, showing the capability of our techniques on large real-world graph datasets. We conclude that \algo and \algog can achieve excellent performance in \production setting with \inductive \& \transductive.

\subsection{Inference Time}\label{sec:time}

\begin{figure}[!t]
\begin{minipage}[t]{.40\linewidth}
    \centering
    \includegraphics[width=\linewidth]{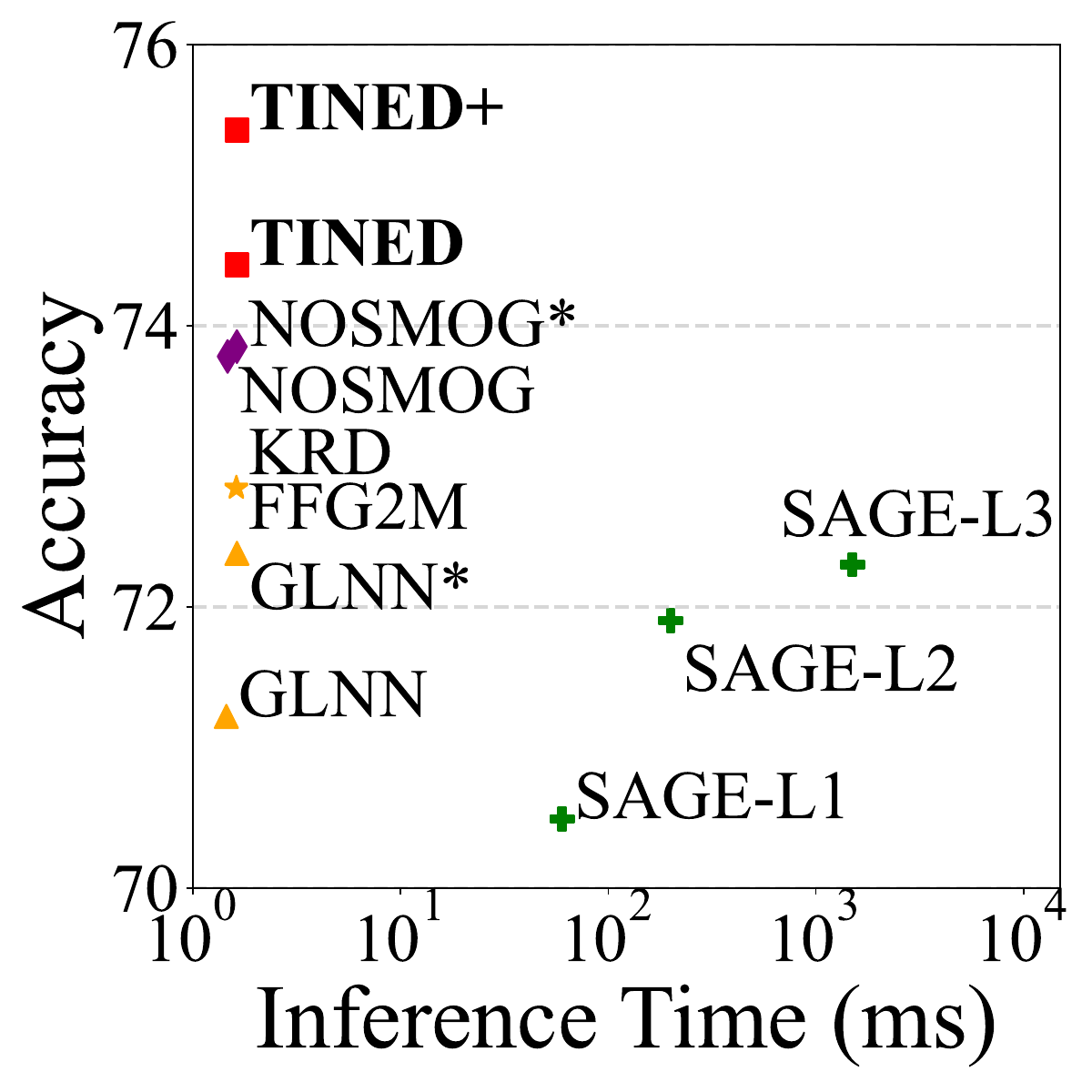}
    \vspace{-4mm}
    \captionof{figure}{{Inference Time and Accuracy}}\label{fig:time}
  \end{minipage}\hfill\hspace{4pt}
  \begin{minipage}[t]{.58\linewidth}
    \centering
\includegraphics[width=1.0\linewidth]{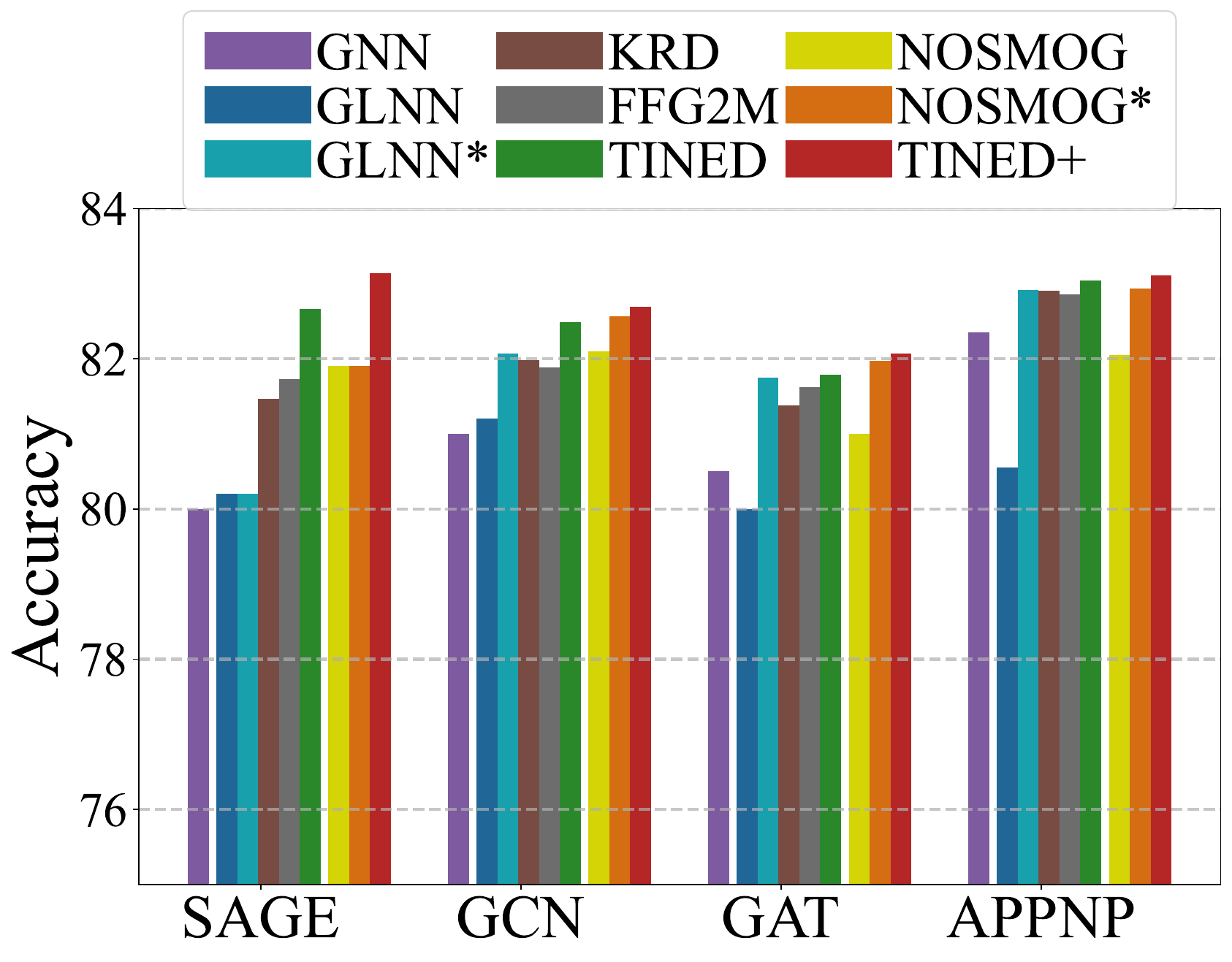}
    \vspace{-4mm}
    \captionof{figure}{{Accuracy on Different Teacher GNNs}}\label{fig:teacherGNNs}
  \end{minipage}

 \begin{minipage}[b]{0.45\linewidth}

 \end{minipage}

  \vspace{-4mm}
\end{figure}

Figure \ref{fig:time} shows the trade-off between accuracy and inference time on \citseer. Methods closer to the top-left corner achieve a better balance of accuracy and speed. Our methods, \algo and \algog, achieve the highest accuracy (74.43\% and 75.39\%) while being fast (1.63ms and 1.64ms). In contrast, GNNs are slower, e.g., 2-layer {\small GraphSAGE (SAGE-L2)} takes 153.14ms, and 3-layer {\small GraphSAGE (SAGE-L3)} takes 1202.45ms. \algo and \algog are 94 times faster than {\small SAGE-L2} and 733 times faster than {\small SAGE-L3}. 
All the distillation methods (\nosmog, \glnnstar, \algo, \algog, KRD and FFG2M) have similar efficiency in around 1-2ms, 
with negligible differences. Compared with SAGE teacher that needs 153.14ms, the efficiency of distillation techniques in our design is validated.

 \begin{figure*}[!t]
 \begin{minipage}[h]{0.64\textwidth}
    \centering
\begin{minipage}[b]{0.19\linewidth}
  \centering
  \includegraphics[width=\linewidth]{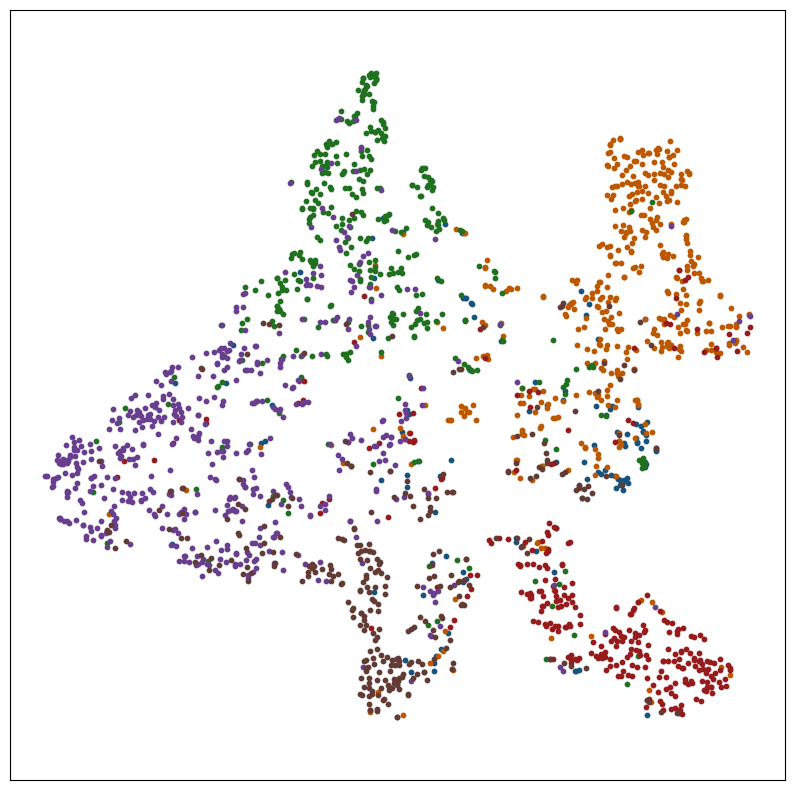}
  \par\vspace{-3pt}
  {\footnotesize (a) SAGE (epoch=500)}
\end{minipage}
\hfill
\begin{minipage}[b]{0.19\linewidth}
  \centering
  \includegraphics[width=\linewidth]{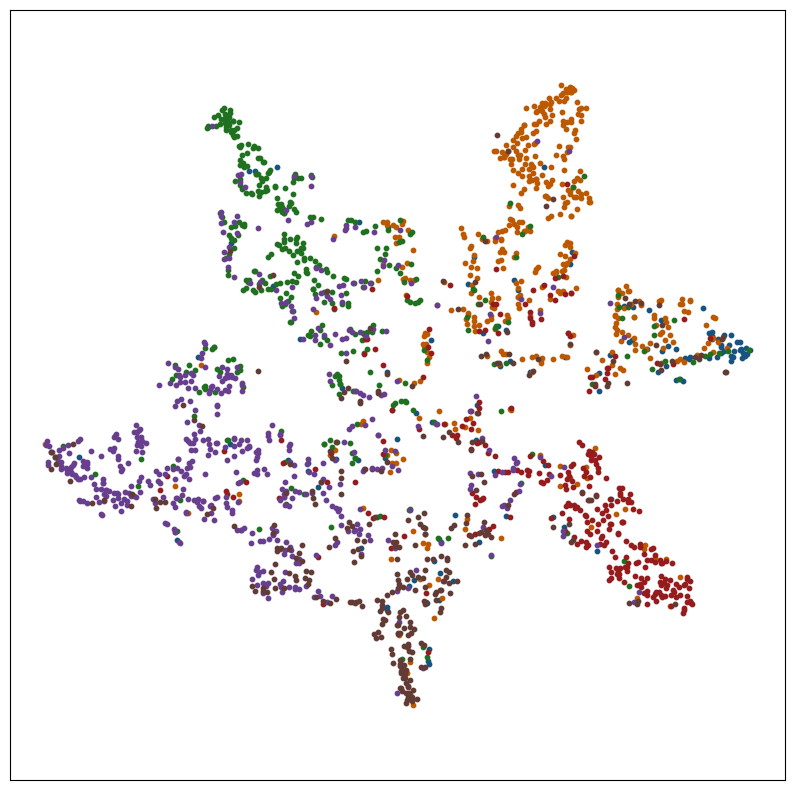}
  \par\vspace{-3pt}
  {\footnotesize (b) \algo (epoch=0)}
\end{minipage}
\hfill
\begin{minipage}[b]{0.19\linewidth}
  \centering
  \includegraphics[width=\linewidth]{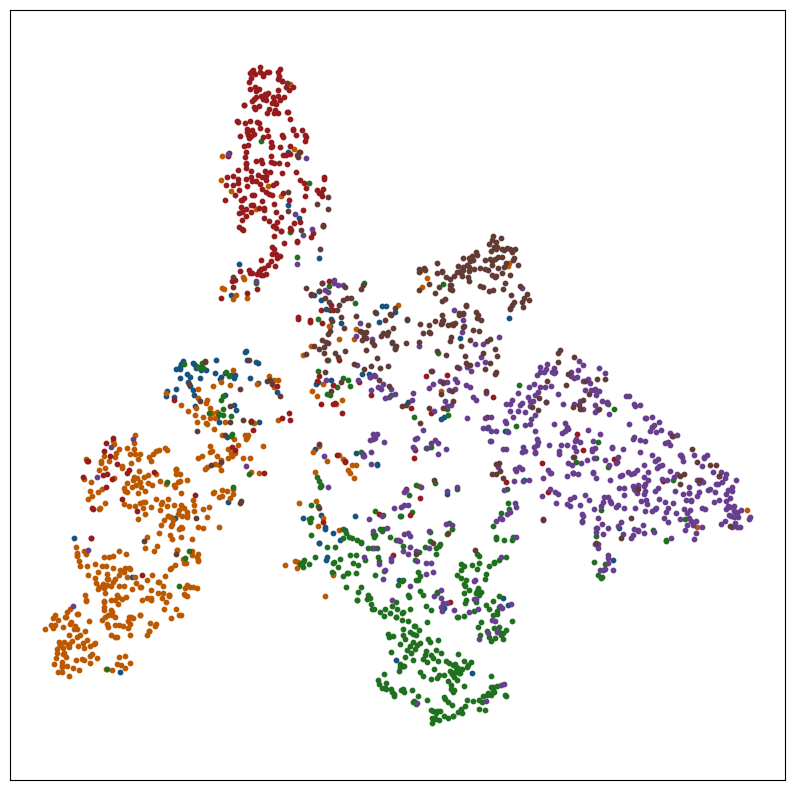}
  \par\vspace{-3pt}
  {\footnotesize (c) {\algo} (epoch=500)}
\end{minipage}
\hfill
\begin{minipage}[b]{0.19\linewidth}
  \centering
  \includegraphics[width=\linewidth]{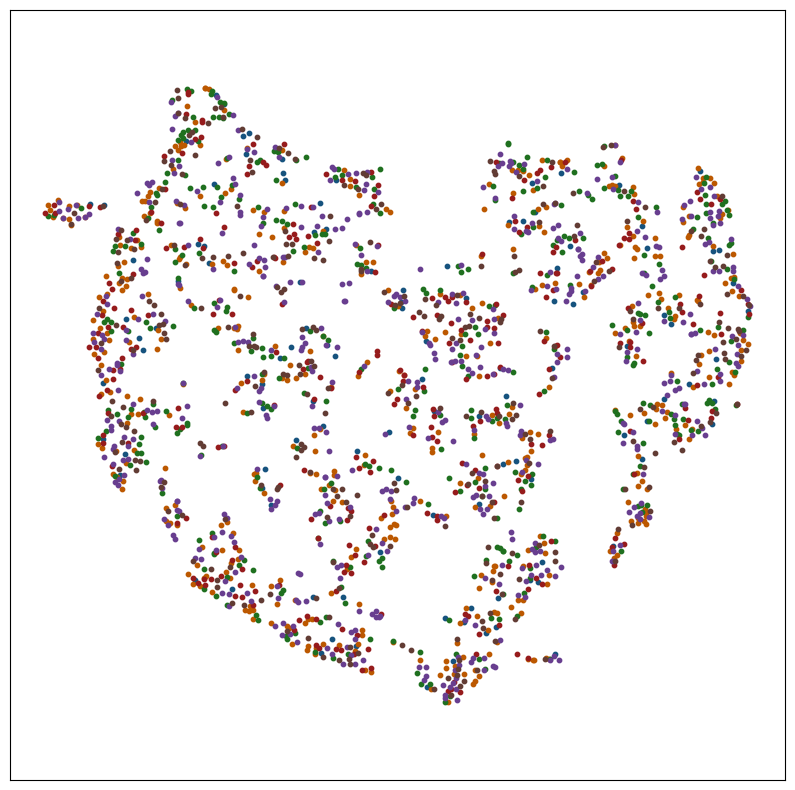}
  \par\vspace{-3pt}
  {\footnotesize (d) GLNN* (epoch=0)}
\end{minipage}
\hfill
\begin{minipage}[b]{0.19\linewidth}
  \centering
  \includegraphics[width=\linewidth]{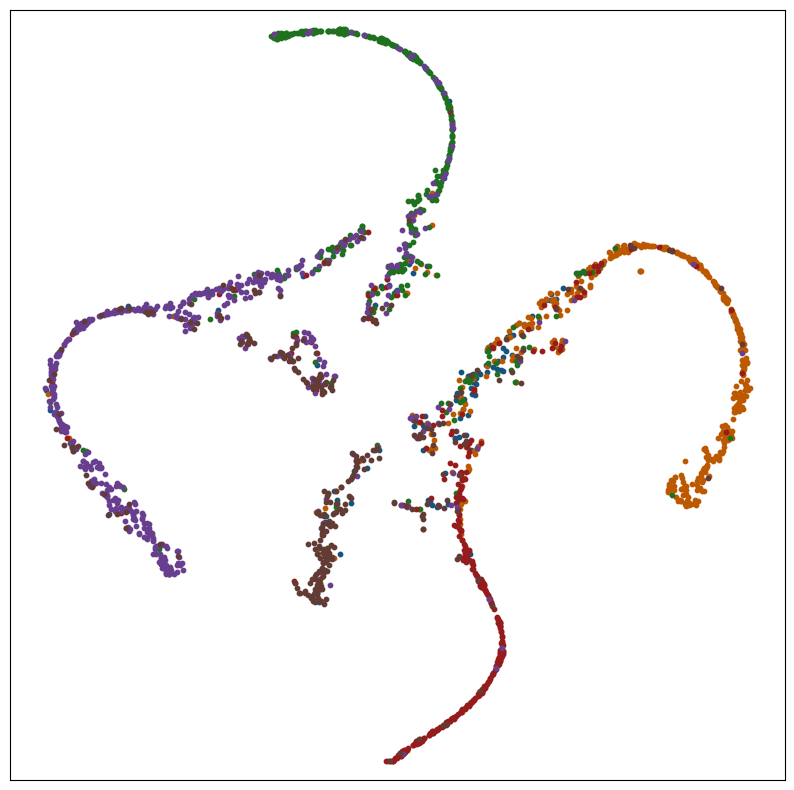}
  \par\vspace{-3pt}
  {\footnotesize (e) \glnnstar (epoch=500)}
\end{minipage}

\vspace{-0mm}
\caption{
  t-SNE of model embeddings at different training stages on Citeseer. 
}
\label{fig:tsne_part}
  \end{minipage} \hfill\hspace{5pt}
\begin{minipage}[h]{.32\textwidth}
\vspace{-3mm}
    \captionof{table}{
Approximation error
}
\vspace{-0mm}
\renewcommand{\arraystretch}{0.9}
\begin{adjustbox}{width=\linewidth,center}
\setlength{\tabcolsep}{2pt}
\begin{tabular}{l|cc|cc}
\toprule
 & \multicolumn{2}{c|}{$l=1$}& \multicolumn{2}{c}{$l=2$}\\
    \midrule
    Datasets & \algo& \glnnstar& \algo&\glnnstar\\ \midrule
    \cora & 1.32& 1.64& 0.90&1.12\\ 
    \citseer & 0.46& 0.86& 0.26&0.75\\
    \pubmed & 0.88& 0.82& 0.74&1.02\\
    \acomputer & 1.89& 1.88& 2.55&2.65\\
    \aphotos & 1.21&  0.83& 1.40&1.54\\
    \bottomrule
    \end{tabular}
\end{adjustbox}
\label{tab:approximationError}
\vspace{-3mm}
  \end{minipage}
  \vspace{0mm}
\end{figure*}

 \begin{figure*}[!t]
\begin{minipage}[h]{0.58\textwidth}
    \centering
		\includegraphics[width=0.95\linewidth]{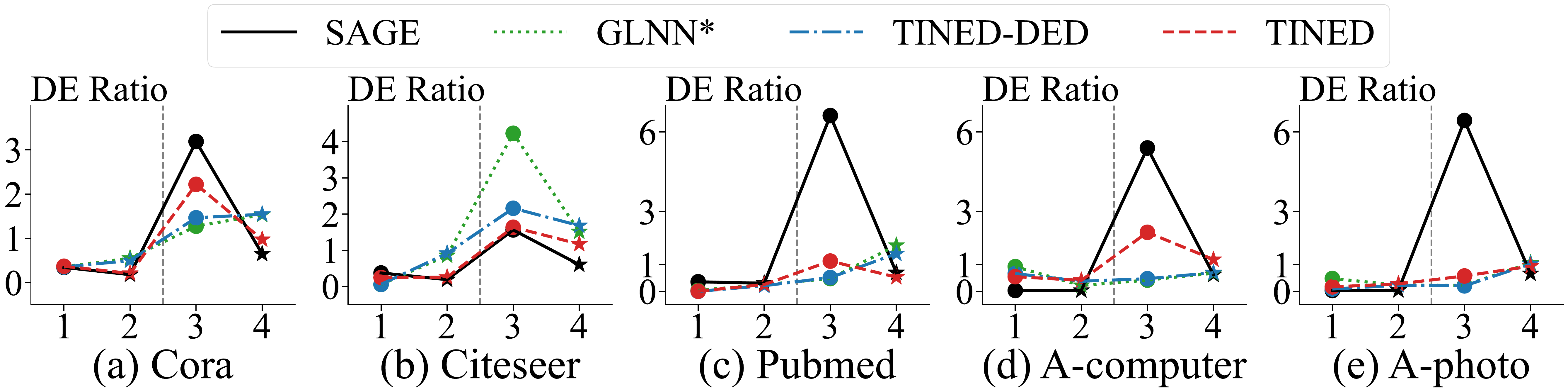}
   \vspace{-0mm}
\caption{Learned DE ratios in MLPs v.s. ground truth  of $2$-layer GraphSAGE. 
}
\label{fig:2_layerwise_DE_ratio_compare}
  \end{minipage}\hfill\hspace{4pt}
  \begin{minipage}[h]{0.41\textwidth}
  \vspace{-6mm}
      \captionof{table}{
    Ablation study
    }   \label{tab:ablation}
\vspace{-0mm}
        \begin{adjustbox}{width=0.98\linewidth,center}
        \setlength{\tabcolsep}{2pt}
        \begin{tabular}{llllll}
            \toprule
            \multicolumn{1}{l}{Datasets}  & \algoNoIn& \algoNoDED & \algo &$\Delta_{\mathsf{TIN}}$& $\Delta_{\mathsf{DED}}$\\ \midrule
            \cora & 81.32$\pm$1.60& 81.64$\pm$1.64& 82.63$\pm$1.57&$\uparrow$ 1.31\%& $\uparrow$ 0.99\%\\ 
            \citseer & 71.88$\pm$1.27& 74.12$\pm$1.42& 74.43$\pm$1.53&$\uparrow$ 2.55\%& $\uparrow$ 0.31\%\\
            \pubmed & 75.93$\pm$3.40& 76.45$\pm$2.68& 77.09$\pm$2.14&$\uparrow$ 1.16\%& $\uparrow$ 0.64\%\\
            \acomputer & 84.59$\pm$1.78& 84.80$\pm$1.67& 85.18$\pm$1.12&$\uparrow$ 0.59\%& $\uparrow$ 0.38\%\\
            \aphotos & 93.90$\pm$0.52& 93.67$\pm$0.69& 93.97$\pm$0.53&$\uparrow$ 0.07\%& $\uparrow$ 0.30\%\\
            \ogba & 63.72$\pm$0.89& 64.39$\pm$0.80& 64.44$\pm$0.72&$\uparrow$ 0.72\%& $\uparrow$ 0.05\%\\
            \ogbp & 69.20$\pm$0.25& 68.63$\pm$0.26& 69.48$\pm$0.25&$\uparrow$ 0.28\%& $\uparrow$ 0.85\%\\
            \bottomrule
            \end{tabular}
        \end{adjustbox}
  \vspace{-3mm}
    \end{minipage}

  \eat{\begin{minipage}[b]{.36\textwidth}
\begin{minipage}[b]{\linewidth}
  \centering
\captionof{table}{Varying $\eta$ on \acomputer.}
\label{tab:eta_results}
\vspace{-2mm}
\begin{adjustbox}{width=\linewidth,center}
\small
\setlength{\tabcolsep}{2pt}
\begin{tabular}{c|c|c|c|c|c|c|c}  
\toprule
$\eta$ & 0 & 1e-9 & 1e-6 & 1e-3 & 1e-1 & 1 & 10 \\ \hline
\algo & 71.99 & 72.52 & 72.72 & 85.18 & 83.60 & 82.84 & 70.62 \\
\bottomrule
\end{tabular}
\vspace{3mm}
\end{adjustbox}
\end{minipage}
\hfill

\begin{minipage}[b]{\linewidth}
  \centering
\captionof{table}{Varying $\beta$ on \cora.}
\label{tab:beta_results}
\vspace{-3mm}
\begin{adjustbox}{width=\linewidth,center}
\small
\setlength{\tabcolsep}{2pt}
\begin{tabular}{c|c|c|c|c|c|c|c}  
\toprule
$\beta$ & 0    & 1e-9 & 1e-6 & 1e-3 & 1e-1 & 1    & 10   \\ \hline
\algo  & 81.54 & 81.32 & 81.64 & 81.68 & 81.70 & 82.61 & 80.92 \\
\bottomrule
\end{tabular}
\end{adjustbox}
\end{minipage}
  \end{minipage}
  }

  \vspace{-3mm}
\end{figure*}

\subsection{Different Teacher GNN Architectures}\label{sec:different}
\vspace{-1mm}

 Figure \ref{fig:teacherGNNs}, we show that \algo and \algog can maintain strong performance with alternative GNN teachers, including \gcn, \gat, and \appnp, in addition to \sage. We report the average performance of a distillation method  with different teacher GNNs across the five benchmark datasets. 
 With different teachers, \algo can always learn more effective student MLPs than  \glnn, \glnnstar, KRD, and FFG2M. With graph dependency, \algog can maintain its strong performance with different teachers and surpasses \nosmog and \nosmogstar.
When different teacher GNNs are adopted, Figure \ref{fig:teacherGNNs} illustrates the effectiveness of our methods to  preserve layer-level knowledge into MLPs.

\subsection{Experimental Analysis}\label{sec:ablation}

\textbf{Ablation Study.} 
Denote \algoNoIn as  \algo without Teacher Injection (TIN), and \algoNoDED as \algo without Dirichlet Energy Distillation (DED)
The results are in Table~\ref{tab:ablation}. 
Both  TIN and DED contribute positively, and 
\algo achieves the best performance on all datasets, demonstrating   the effectiveness of our techniques in \algo.

\eat{\begin{table}[t]
\centering
\caption{
Ablation Study
}
\vspace{-3mm}
    \renewcommand{\arraystretch}{0.94}
\begin{adjustbox}{width=0.98\linewidth,center}
\setlength{\tabcolsep}{2pt}
\begin{tabular}{llllll}
    \toprule
    \multicolumn{1}{l}{Datasets}  & \algoNoIn& \algoNoDED & \algo &$\Delta_{\mathsf{TIN}}$& $\Delta_{\mathsf{DED}}$\\ \midrule
    \cora & 81.32$\pm$1.60& 81.64$\pm$1.64& 82.63$\pm$1.57&$\uparrow$ 1.31\%& $\uparrow$ 0.99\%\\ 
    \citseer & 71.88$\pm$1.27& 74.12$\pm$1.42& 74.43$\pm$1.53&$\uparrow$ 2.55\%& $\uparrow$ 0.31\%\\
    \pubmed & 75.93$\pm$3.40& 76.45$\pm$2.68& 77.09$\pm$2.14&$\uparrow$ 1.16\%& $\uparrow$ 0.64\%\\
    \acomputer & 84.59$\pm$1.78& 84.80$\pm$1.67& 85.18$\pm$1.12&$\uparrow$ 0.59\%& $\uparrow$ 0.38\%\\
    \aphotos & 93.90$\pm$0.52& 93.67$\pm$0.69& 93.97$\pm$0.53&$\uparrow$ 0.07\%& $\uparrow$ 0.30\%\\
    \ogba & 63.72$\pm$0.89& 64.39$\pm$0.80& 64.44$\pm$0.72&$\uparrow$ 0.72\%& $\uparrow$ 0.05\%\\
    \ogbp & 69.20$\pm$0.25& 68.63$\pm$0.26& 69.48$\pm$0.25&$\uparrow$ 0.28\%& $\uparrow$ 0.85\%\\
    \bottomrule
    \end{tabular}
\end{adjustbox}
\label{tab:ablation}
\vspace{-3mm}
\end{table} 
}

\eat{\begin{figure}[!t]
	\centering
		\includegraphics[width=1.04\columnwidth]{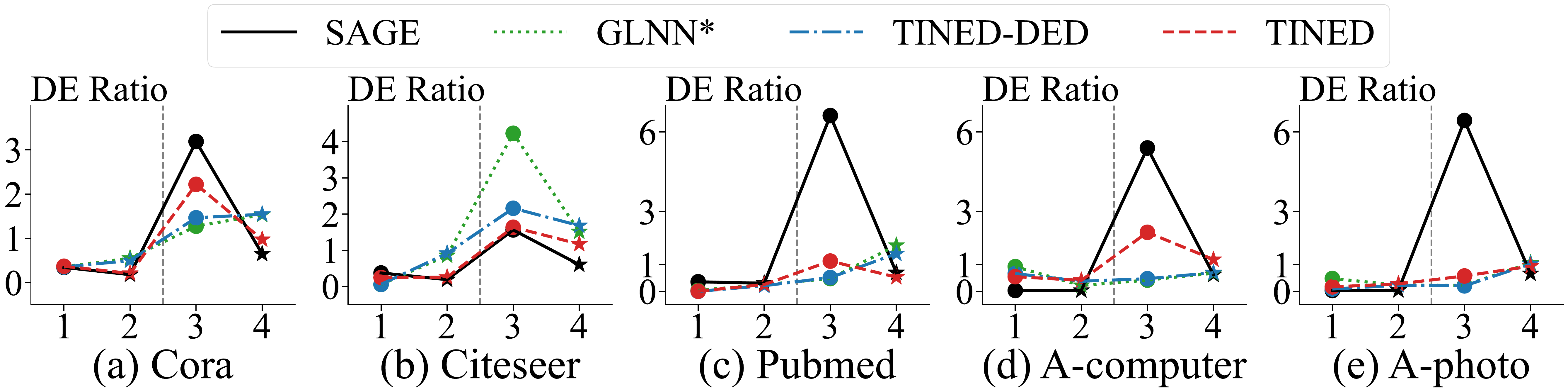}
   \vspace{-4mm}
\caption{Learned DE ratios in MLPs v.s. ground truth  of $2$-layer GraphSAGE. 
}
\label{fig:2_layerwise_DE_ratio_compare}
\vspace{-3mm}
\end{figure}}

\textbf{Visualization.} 
Based on the visualization of teacher SAGE's output embeddings after 500 epochs of training (Figure \ref{fig:tsne_part}(a)), we compare the embeddings of \algo and \glnnstar both at initialization (epoch 0) and after training convergence (epoch 500).
Figure \ref{fig:tsne_part}(b) shows \algo initialized (epoch 0) using the proposed teacher injection technique, resulting in an embedding visualization similar to that of the teacher in Figure \ref{fig:tsne_part}(a). In contrast, the visualization of \glnnstar at epoch 0 appears quite random, as seen in Figure \ref{fig:tsne_part}(d).
Upon reaching training convergence (epoch 500), \algo in Figure \ref{fig:tsne_part}(c) maintains a visualization closely resembling the teacher's in Figure \ref{fig:tsne_part}(a), whereas the baseline \glnnstar displays a noticeably different visualization. This evaluation confirms the effectiveness of the proposed teacher injection and Dirichlet energy distillation techniques employed by \algo. More visualization results are in \ref{sec:tsne_all}.

\begin{table}[t]
  \centering
  \caption{Vary $\eta$ of \algo on transductive setting. Best results are in bold.}
  \label{tab:vary_eta_qs85}
    
\begin{adjustbox}{width=1\linewidth,center}
  \begin{tabular}{@{}llllll@{}}
    \toprule
    $\eta$ & \cora & \citseer & \pubmed & \acomputer & \aphotos \\
    \midrule
    1e-09 & 73.80$\pm$2.56 & 67.20$\pm$1.86 & 73.39$\pm$2.10 & 72.53$\pm$2.17 & 86.06$\pm$2.38 \\
    1e-06 & 73.50$\pm$2.48 & 67.19$\pm$1.87 & 73.32$\pm$1.88 & 72.73$\pm$1.99 & 86.35$\pm$1.79 \\
    0.001 & 77.12$\pm$0.91 & 67.24$\pm$1.83 & 73.69$\pm$1.83 & 84.13$\pm$1.23 & 87.30$\pm$2.08 \\
    0.01 & 79.26$\pm$1.30 & 69.38$\pm$1.70 & 75.12$\pm$2.32 & \textbf{85.17$\pm$1.21} & 90.48$\pm$1.38 \\
    0.1 & 81.40$\pm$1.69 & 71.52$\pm$1.58 & 76.55$\pm$2.81 & 83.61$\pm$1.67 & \textbf{93.97$\pm$0.54} \\
    0.5 & 82.01$\pm$1.64 & \textbf{74.57$\pm$1.42} & \textbf{77.10$\pm$2.15} & 83.23$\pm$1.29 & 93.51$\pm$0.60 \\
    1.0 & \textbf{82.61$\pm$1.58} & 73.57$\pm$1.39 & 76.65$\pm$2.77 & 82.85$\pm$0.91 & 93.37$\pm$0.51 \\
    10.0 & 78.85$\pm$1.63 & 73.56$\pm$1.48 & 75.61$\pm$2.73 & 70.63$\pm$4.87 & 89.53$\pm$0.78 \\
    \bottomrule
  \end{tabular}
  \end{adjustbox}

\end{table}

\noindent\textbf{DE ratio.} 
Recall that Figure \ref{fig:2_layerwise_DE_ratio_sage} shows the ground-truth DE ratios of teacher \sage, and we propose Dirichlet Energy Distillation in Section \ref{sec:ded}  to preserve smoothing effects into MLPs. 
Figure \ref{fig:2_layerwise_DE_ratio_compare} reports the learned DE ratios of \algo, \algoNoDED, and \glnnstar. \algo is more closely aligned to the ground-truth DE ratios, highlighting that \algo effectively retains the GNN smoothing properties into MLPs.

\eat{
\begin{table}[!t]
\centering
\caption{
Approximation error
}
\vspace{-3mm}
\renewcommand{\arraystretch}{0.86}
\begin{adjustbox}{width=0.66\linewidth,center}
\setlength{\tabcolsep}{2pt}
\begin{tabular}{l|cc|cc}
\toprule
 & \multicolumn{2}{c|}{$l=1$}& \multicolumn{2}{c}{$l=2$}\\
    \midrule
    Datasets & \algo& \glnnstar& \algo&\glnnstar\\ \midrule
    \cora & 1.32& 1.64& 0.90&1.12\\ 
    \citseer & 0.46& 0.86& 0.26&0.75\\
    \pubmed & 0.88& 0.82& 0.74&1.02\\
    \acomputer & 1.89& 1.88& 2.55&2.65\\
    \aphotos & 1.21&  0.83& 1.40&1.54\\
    \bottomrule
    \end{tabular}
\end{adjustbox}
\label{tab:approximationError}
\vspace{-3mm}
\end{table} 
}

\noindent\textbf{Approximation Bound.}
Theorem \ref{theoremErrorBound} shows a bound between $\gp^\la$ in GNNs and its counterpart $\fc^\laone$ in MLPs.
Table \ref{tab:approximationError} reports the errors of \algo and \glnnstar using $\fc^\laone$ to approximate $\gp^\la$, $ ||  \gp^\la (\bH)- \fc^{\laone} (\bH)  ||_F   /||\bH||_F$, for $l=1,2$.
When $l=2$, the errors of \algo are  lower than \glnnstar, while the errors for $l=1$ are comparable, showing \algo effectively approximates $\gp^\la$.

\eat{
\begin{figure*}[t]
\centering
\begin{minipage}[t]{0.18\textwidth}
  \centering
  \includegraphics[width=\linewidth]{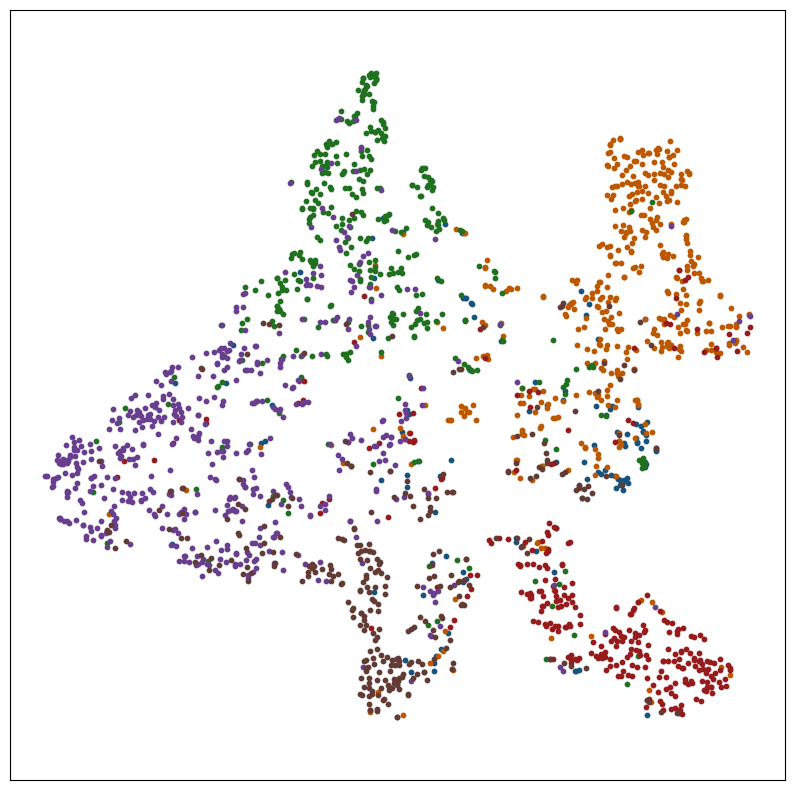}
  \par\vspace{-3pt}
  \footnotesize (a) SAGE (epoch=500)
\end{minipage}
\hfill
\begin{minipage}[t]{0.18\textwidth}
  \centering
  \includegraphics[width=\linewidth]{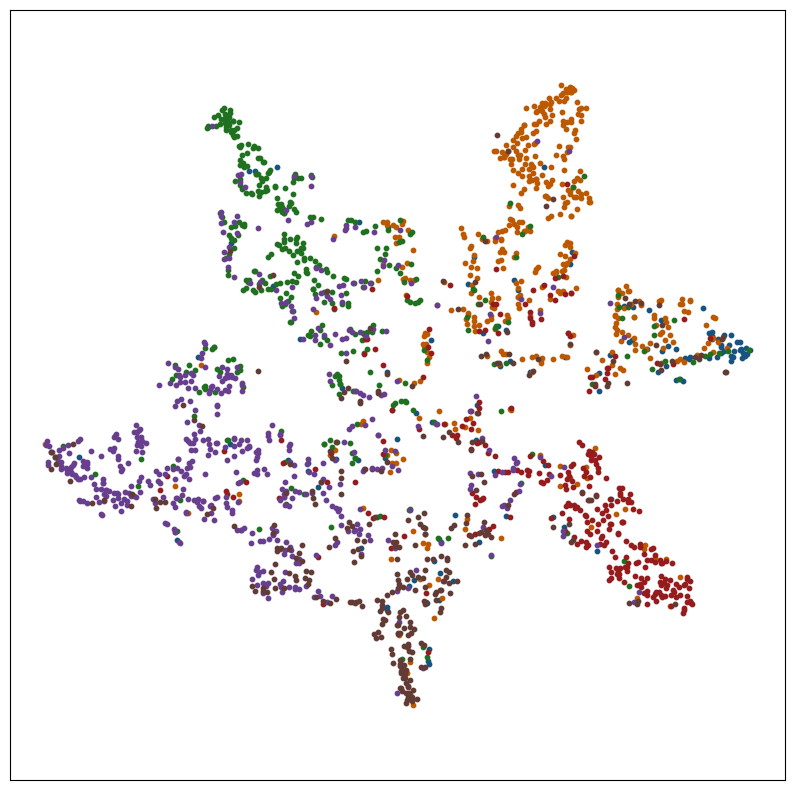}
  \par\vspace{-3pt}
  \footnotesize (b) \algo (epoch=0)
\end{minipage}
\hfill
\begin{minipage}[t]{0.18\textwidth}
  \centering
  \includegraphics[width=\linewidth]{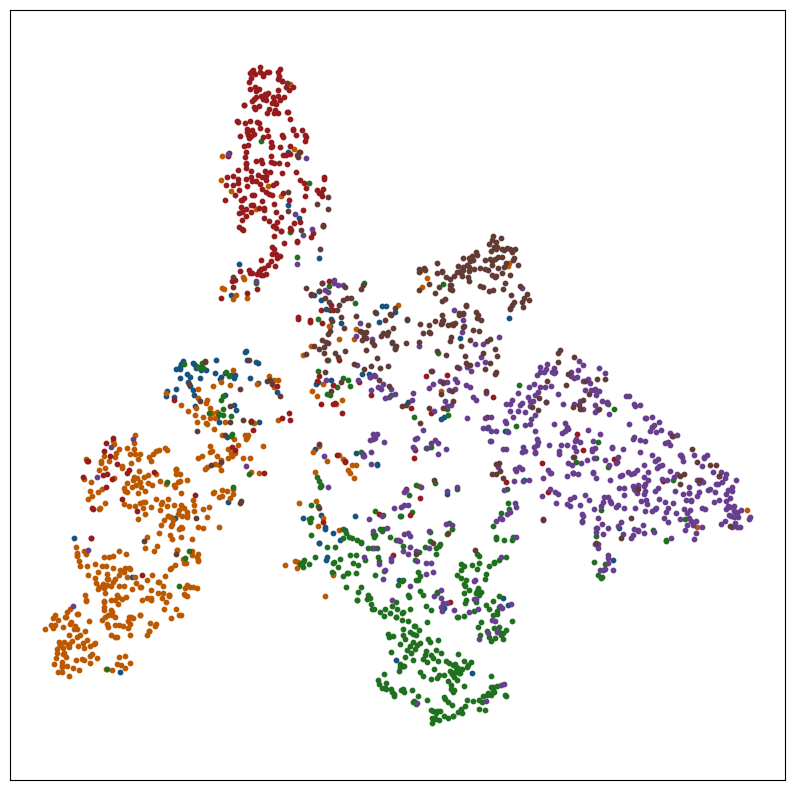}
  \par\vspace{-3pt}
  \footnotesize (c) \algo (epoch=500)
\end{minipage}
\hfill
\begin{minipage}[t]{0.18\textwidth}
  \centering
  \includegraphics[width=\linewidth]{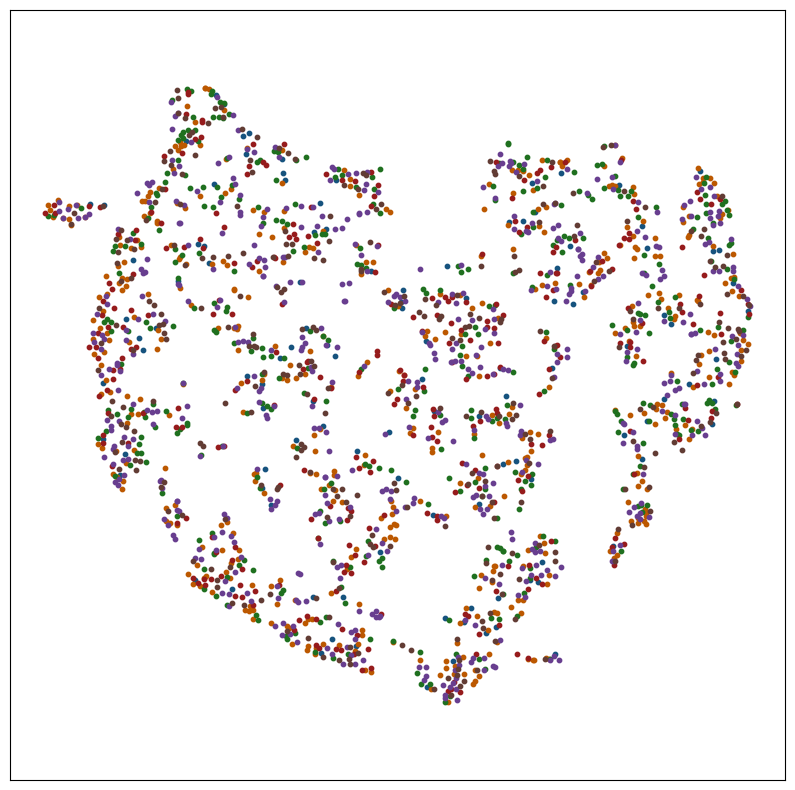}
  \par\vspace{-3pt}
  \footnotesize (d) GLNN* (epoch=0)
\end{minipage}
\hfill
\begin{minipage}[t]{0.18\textwidth}
  \centering
  \includegraphics[width=\linewidth]{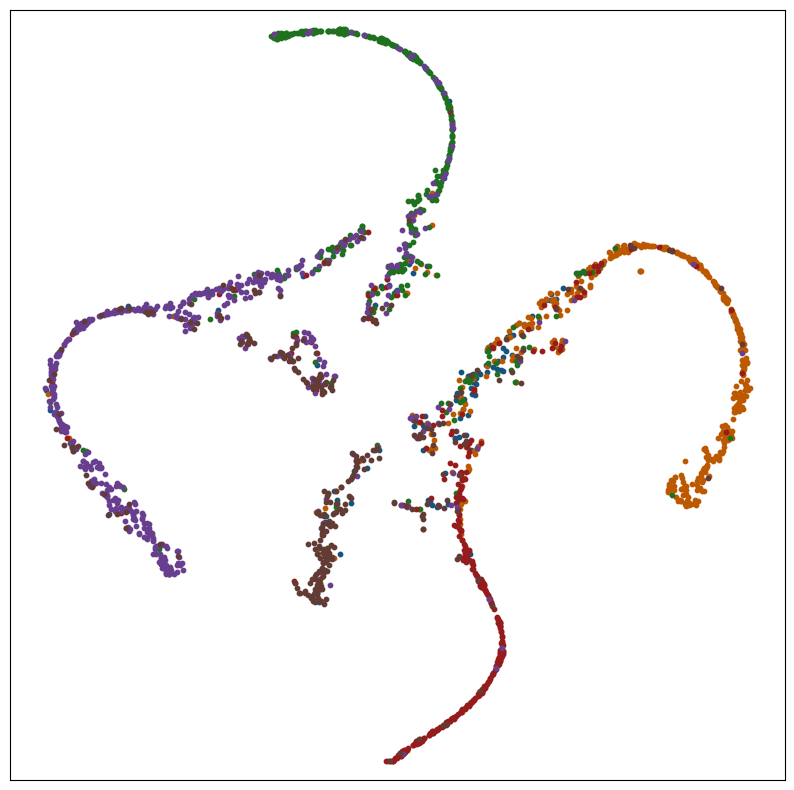}
  \par\vspace{-3pt}
  \footnotesize (e) \glnnstar (epoch=500)
\end{minipage}

\vspace{-2mm}
\caption{
  t-SNE visualization of the output embeddings of models at different training stages on Citeseer.

}
\label{fig:tsne_part}
\end{figure*}
}

\eat{\begin{table}[!t]
\centering
\caption{Varying $\eta$ on computer.}
\label{tab:eta_results}
\vspace{-3mm}
\begin{adjustbox}{width=0.78\linewidth,center}
\small
\setlength{\tabcolsep}{3pt}
\begin{tabular}{c|c|c|c|c|c|c|c}  
\toprule
$\eta$ & 0 & 1e-9 & 1e-6 & 1e-3 & 1e-1 & 1 & 10 \\ \hline
\algo & 71.99 & 72.52 & 72.72 & 85.18 & 83.60 & 82.84 & 70.62 \\
\bottomrule
\end{tabular}
\end{adjustbox}
\vspace{-3mm}

\end{table}}

\eat{\begin{table}[!t]
\centering
\caption{Vary $\beta$ on Cora.}\label{tab:beta_results}
\vspace{-3mm}
\begin{adjustbox}{width=0.78\linewidth,center}
\small
\setlength{\tabcolsep}{3pt}
\begin{tabular}{c|c|c|c|c|c|c|c}  
\toprule
$\beta$ & 0    & 1e-9 & 1e-6 & 1e-3 & 1e-1 & 1    & 10   \\ \hline
\algo  & 81.54 & 81.32 & 81.64 & 81.68 & 81.70 & 82.61 & 80.92 \\
\bottomrule
\end{tabular}
\end{adjustbox}
\vspace{-3mm}

\end{table}}

\begin{table*}[t]
    \centering
    \caption{
 Experiment result under \textit{prod} (\textit{ind} \& \textit{tran}) setting on heterophilic datasets. 
  The best result in each category is in bold. 
    }
    \vspace{-2mm}
    \setlength{\tabcolsep}{3pt}
    \renewcommand{\arraystretch}{0.92}

   \begin{adjustbox}{width=0.98\textwidth,center}
   
    \begin{tabular}{ll|l|lllll|lll}
    \toprule
 & & Teacher& \multicolumn{5}{c|}{Without Graph Dependency}& \multicolumn{3}{c}{With Graph Dependency}\\
    \midrule
    
    {Datasets} & Eval & SAGE &  {FFG2M} & {KRD} & GLNN  &\glnnstar&
\algo & NOSMOG &NOSMOG*& \algog\\ \midrule

    \multirow{3}{*}{Squirrel} & \production & 35.47 &38.34 &37.16& 39.90 & 39.70 & \textbf{41.95} & 38.17 & 39.43 & \textbf{40.89}\\ 
            & \inductive &  41.44$\pm$4.66 &42.00$\pm$4.78 &42.11$\pm$4.77& 44.89$\pm$5.67 & 45.00$\pm$4.56 & \textbf{46.89$\pm$5.23} & 45.44$\pm$4.75 & 44.33$\pm$3.24 & \textbf{46.78$\pm$3.80}\\
            & \transductive &33.98$\pm$1.66 &37.43$\pm$3.34 &35.93$\pm$3.36& 38.65$\pm$1.09 & 38.37$\pm$1.02 & \textbf{40.72$\pm$1.36}&  36.35$\pm$1.69 &38.20$\pm$1.59 & \textbf{39.42$\pm$1.05}  \\ \midrule
    \multirow{3}{*}{Amazon-ratings} & \production &  47.55 & 50.33 &49.56& 49.87 & 49.41 & \textbf{50.70}& 47.86 & 48.80 & \textbf{50.42}  \\ 
            & \inductive & 47.45$\pm$1.48 &47.55$\pm$0.97 &47.68$\pm$1.00& 47.72$\pm$1.00 & 47.71$\pm$1.14 & \textbf{49.02$\pm$1.02}&47.47$\pm$1.45 & 48.46$\pm$1.37 & \textbf{49.51$\pm$1.54}  \\
            & \transductive &47.58$\pm$0.48  & 51.03$\pm$1.73   &50.03$\pm$1.50& 50.41$\pm$0.45 & 49.84$\pm$0.39 &\textbf{51.12$\pm$0.44}  & 47.96$\pm$0.31 & 48.88$\pm$0.56 & \textbf{50.65$\pm$0.56} \\ 
    \bottomrule
    \end{tabular}
    \end{adjustbox}
    \label{tab:prod_heterophily}
    \vspace{-2mm}
\end{table*}

\begin{table}[t]
  \centering
  \caption{Vary $\beta$ of \algo on transductive setting. Best results are in bold. }
  \label{tab:vary_beta_qs85}

\begin{adjustbox}{width=1\linewidth,center}
  \setlength{\tabcolsep}{3pt}
  \begin{tabular}{@{}llllll@{}}
    \toprule
    $\beta$ & \cora & \citseer & \pubmed & \acomputer & \aphotos \\
    \midrule
    1e-09 & 81.33$\pm$1.49 & 73.39$\pm$1.28 & 76.63$\pm$2.42 & 84.70$\pm$1.15 & 93.41$\pm$0.61 \\
    1e-06 & 81.64$\pm$1.71 & 73.39$\pm$1.30 & 76.28$\pm$2.76 & \textbf{85.17$\pm$1.21} & 93.48$\pm$0.65 \\
    0.001 & 81.64$\pm$1.57 & 73.39$\pm$1.31 & 76.51$\pm$2.64 & 84.80$\pm$1.21 & 93.65$\pm$0.68 \\
    0.1 & 81.71$\pm$1.59 & 73.65$\pm$1.38 & \textbf{77.10$\pm$2.15} & 84.70$\pm$0.99 & \textbf{93.97$\pm$0.58} \\
    1.0 & \textbf{82.61$\pm$1.58} & 73.81$\pm$1.29 & 75.84$\pm$2.66 & 84.29$\pm$1.08 & 86.59$\pm$5.66 \\
    10.0 & 80.92$\pm$2.15 & \textbf{74.57$\pm$1.42} & 71.57$\pm$2.76 & 71.35$\pm$6.30 & 78.00$\pm$3.19 \\
    \bottomrule
  \end{tabular}
  \end{adjustbox}
 \end{table}

\textbf{Parameter Sensitivity Analysis.}
In our method, the parameter $\eta$ controls the degree of fine-tuning in Eq.~\ref{eq:fine_tuning}, while $\beta$ controls the importance of Dirichlet Energy Distillation in Eq.~\ref{eq:total_loss}. 
Tables \ref{tab:vary_eta_qs85} and \ref{tab:vary_beta_qs85} present the accuracy results of varying $\eta$ and $\beta$ of \algo across multiple datasets. 
In Table \ref{tab:vary_eta_qs85}, as $\eta$ increases, \algo demonstrates a clear trend where performance initially improves and then declines, with the best results highlighted in bold. 
A similar pattern is observed for $\beta$ in Table \ref{tab:vary_beta_qs85}, emphasizing the trade-off controlled by these parameters.  
Note that hyperparameter tuning is essential in machine learning research. 
Appendix~\ref{app:teacherhyper} of the paper details the search space for our parameters.

\textbf{Heterophilic Datasets}
We conducted experiments on representative heterophilic datasets, Squirrel and Amazon-ratings, under the production setting. 
Squirrel~\cite{squirrel} is a web page dataset collected from Wikipedia, while Amazon-ratings~\cite{platonov2023critical} is a product co-purchasing network based on data from SNAP Datasets.
In Table \ref{tab:prod_heterophily}, our methods, \algo and \algog, consistently outperform existing approaches across all settings, often by a significant margin. 
For instance, on the Squirrel dataset without graph dependency, \algo achieves a \production performance of 41.95\%, surpassing the best competitor \glnn, which achieves 39.90\%. 
Similarly, on the Amazon-ratings dataset with graph dependency, \algog achieves a \production accuracy of 50.42\%, representing a 1.62\% improvement over the best competitor \nosmogstar. 
These results highlight the effectiveness of our proposed methods in handling heterophilic datasets.

\textbf{Comparison with Vanilla MLP with Additional Layers.}
In addition to the vanilla MLP with 2 layers, we also evaluate a 4-layer MLP, denoted as MLP*. Tables \ref{tab:compare_mlp_transductive_jzck} and \ref{tab:compare_mlp_prod_jzck} present the results under both transductive and production settings. Our method, \algo, consistently outperforms MLP* across all datasets.
Interestingly, MLP* demonstrates degraded performance compared to the 2-layer MLP, suggesting that additional layers lead to overfitting on these datasets. For instance, on the \pubmed dataset, MLP* achieves 67.61\%, which is inferior to the 69.41\% achieved by the 2-layer MLP. In contrast, \algo significantly improves upon MLP*, achieving a notable accuracy of 77.09\%, highlighting its robustness and effectiveness.
 
\begin{table}[t]
  \centering
  \caption{Compare with MLP* with 4 layers under transductive setting.}
  \label{tab:compare_mlp_transductive_jzck}
  \begin{adjustbox}{width=1\linewidth,center}
    \setlength{\tabcolsep}{3pt}
  \begin{tabular}{@{}llllll@{}}
    \toprule
    & \cora & \citseer & \pubmed& \acomputer & \aphotos \\
    \midrule
    MLP & 60.84$\pm$1.08 & 63.41$\pm$1.96 & 69.41$\pm$2.88 & 70.07$\pm$1.77 & 80.19$\pm$1.48 \\
    MLP* & 58.20$\pm$3.16 & 60.81$\pm$3.48 & 67.61$\pm$1.51 & 66.84$\pm$3.03 & 78.66$\pm$2.48 \\
    \algo & 82.63$\pm$1.57 & 74.43$\pm$1.53 & 77.09$\pm$2.14 & 85.18$\pm$1.12 & 93.97$\pm$0.53 \\
    \bottomrule
  \end{tabular}
  \end{adjustbox}
    \vspace{-3mm}
\end{table}

\begin{table}[t]
  \centering
  \caption{Comparison with MLP* with 4 layers under \textit{prod}(\textit{ind}\&\textit{tran}) setting}
  \label{tab:compare_mlp_prod_jzck}
  \begin{adjustbox}{width=1\linewidth,center}
    \setlength{\tabcolsep}{2pt}
  \begin{tabular}{@{}lllllll@{}}
    \toprule
    & Eval &\cora & \citseer & \pubmed& \acomputer & \aphotos\\
    \midrule
    MLP & \textit{ind} & 61.31$\pm$2.16 & 63.95$\pm$2.95 & 69.66$\pm$2.68 & 70.36$\pm$2.48 & 79.76$\pm$2.00 \\
    & \textit{tran} & 60.88$\pm$1.41 & 62.99$\pm$2.39 & 69.67$\pm$2.61 & 69.92$\pm$2.03 & 79.53$\pm$2.05 \\
    \midrule
    MLP* & \textit{ind} & 58.67$\pm$2.42 & 62.15$\pm$3.50 & 67.76$\pm$1.88 & 68.09$\pm$2.60 & 77.27$\pm$2.17 \\
    & \textit{tran} & 58.12$\pm$1.81 & 61.46$\pm$2.46 & 68.07$\pm$1.87 & 67.90$\pm$2.41 & 77.05$\pm$2.48 \\
    \midrule
    TINED & \textit{ind} & 74.38$\pm$1.28 & 72.68$\pm$1.97 & 75.64$\pm$3.02 & 82.83$\pm$1.45 & 91.96$\pm$0.72 \\
    & \textit{tran} &  80.04$\pm$1.50 & 72.20$\pm$1.66 & 75.83$\pm$2.81 & 84.87$\pm$1.38 & 93.74$\pm$0.51 \\
    \bottomrule
  \end{tabular}
  \end{adjustbox}
    \vspace{-3mm}
\end{table}

\eat{
\begin{table*}[htbp]
  \centering
  \caption{\textit{prod} (\textit{ind} \& \textit{tran}) setting on heterophilic dataset (without graph dependency)}
  \label{tab:hetero_no_graph_qs85}
  \setlength{\tabcolsep}{3pt}
  \begin{tabular}{@{}llllllll@{}}
    \toprule
    Data & Eval & SAGE & KRD &FFG2M&GLNN & GLNN* & TINED \\
    \midrule
    Squirrel & \textit{prod} & 35.47 &38.34 &37.16& 39.90 & 39.70 & \textbf{41.95} \\
    & \textit{ind} & 41.44$\pm$4.66 &42.00$\pm$4.78 &42.11$\pm$4.77& 44.89$\pm$5.67 & 45.00$\pm$4.56 & \textbf{46.89$\pm$5.23} \\
    & \textit{tran} & 33.98$\pm$1.66 &37.43$\pm$3.34 &35.93$\pm$3.36& 38.65$\pm$1.09 & 38.37$\pm$1.02 & \textbf{40.72$\pm$1.36} \\
    Amazon-ratings & \textit{prod} & 47.55 & 50.33 &49.56& 49.87 & 49.41 & \textbf{50.70} \\
    & \textit{ind} & 47.45$\pm$1.48 &47.55$\pm$0.97 &47.68$\pm$1.00& 47.72$\pm$1.00 & 47.71$\pm$1.14 & \textbf{49.02$\pm$1.02} \\
    & \textit{tran} & 47.58$\pm$0.48  & 51.03$\pm$1.73   &50.03$\pm$1.50& 50.41$\pm$0.45 & 49.84$\pm$0.39 &\textbf{51.12$\pm$0.44} \\
    \bottomrule
  \end{tabular}
\end{table*}

\begin{table}[htbp]
  \centering
  \caption{\textit{prod} (\textit{ind} \& \textit{tran}) setting on heterophilic dataset (with graph dependency)}
  \label{tab:hetero_with_graph_qs85}
  \begin{tabular}{@{}lllll@{}}
    \toprule
    Data & Eval & NOSMOG &NOSMOG* & TINED+ \\
    \midrule
    Squirrel & \textit{prod} & 38.17 & 39.43 & \textbf{40.89} \\
    & \textit{ind} & 45.44$\pm$4.75 & 44.33$\pm$3.24 & \textbf{46.78$\pm$3.80} \\
    & \textit{tran} &  36.35$\pm$1.69 &38.20$\pm$1.59 & \textbf{39.42$\pm$1.05} \\
    Amazon-ratings & \textit{prod} & 47.86 & 48.80 & \textbf{50.42} \\
    & \textit{ind} &47.47$\pm$1.45 & 48.46$\pm$1.37 & \textbf{49.51$\pm$1.54} \\
    & \textit{tran} & 47.96$\pm$0.31 & 48.88$\pm$0.56 & \textbf{50.65$\pm$0.56} \\
    \bottomrule
  \end{tabular}
\end{table}

}

%% file: contents/conclusion.tex
\section{Conclusion
}\label{sec:conclusion}
We introduce \algo, a novel method for distilling GNN knowledge into MLPs. Our approach includes Teacher Injection, which directly transfers well-trained parameters from GNNs to  MLPs, and Dirichlet Energy Distillation, which preserves the unique smoothing effects of key GNN operations within MLPs. Comprehensive experiments demonstrate that \algo outperforms existing methods across various settings and seven datasets. Currently, the number of layers in the student MLP is dependent on the number of layers in the GNN teacher. In future work, we aim to develop new techniques to accelerate the process and allow for a flexible number of fully connected layers by considering the intrinsic properties of GNNs and MLPs. Moreover, we will explore the potential of \algo in other types of grpahs, including heterogeneous graphs and dynamic graphs, to further enhance its applicability and effectiveness.

%% file: contents/appendix.tex
\appendix

\section{Appendix}\label{appendixSec}
\subsection{Datasets} \label{app:data}

In Table \ref{tab:dataset}, we provide the statistics of the datasets in experiments.
For all datasets, we follow the setting in \cite{glnn, cpf_plp_ft} to split the data. Specifically, for the first five datasets, we use the splitting in \cite{cpf_plp_ft} and each random seed corresponds to a different split. For the OGB datasets \ogba{} and \ogbp{}, we follow the OGB official splits based on time and popularity respectively.

\begin{table*}[!h]
\center
\caption{Dataset Statistics and Splits.}
\small
\begin{tabular}{@{}llllllll}
\toprule
{Dataset} & {\# Nodes} & \# Edges & \# Features & \# Classes  & \# Train& \# Val&\# Test\\ \midrule
\cora & 2,485 & 5,069  & 1,433 & 7   & 140&  210& 2135\\
\citseer & 2,110 & 3,668 & 3,703 & 6    & 120&  180& 1810\\ 
\pubmed & 19,717 & 44,324 & 500 & 3    & 60&  90& 19567\\ 
\acomputer & 13,381 & 245,778  & 767 & 10   & 200&  300& 12881\\
\aphotos & 7,487 & 119,043 & 745 & 8  &  160&  240& 7087\\
\ogba & 169,343 & 1,166,243 & 128 & 40  & 90941&  29799 &48603\\
\ogbp & 2,449,029 & 61,859,140 & 100 & 47 & 196615 & 39323 &2213091\\ 
\bottomrule
\end{tabular}
\label{tab:dataset}
\end{table*}

\subsection{Proof of Theorem~\ref{theoremErrorBound}}\label{proof:errorbound}

\begin{proof}
    We denote the $\mathbf{H}^{\dagger}$ as the pseudo-inverse of matrix $\mathbf{H}$. Since $\mathbf{rank}(\mathbf{H})=d$, $\mathbf{H}$ has linearly independent columns, its pseudo-inverse could be represented as $\mathbf{H}^{\dagger}=(\mathbf{H}^{\top}\mathbf{H})^{-1}\mathbf{H}^{\top}$. 
    Moreover, given the SVD factorization of matrix $\mathbf{H}=\mathbf{U}\left(\begin{array}{c}
         \mathbf{S}  \\
         0 
    \end{array}\right)\mathbf{V}^{\top}$, where $\mathbf{U}\in\mathbb{R}^{N\times N}$ and $\mathbf{V}\in\mathbb{R}^{d\times d}$ are unitary, and $\mathbf{S}\in\mathbb{R}^{d\times d}$ is invertible diagonal matrix,  we have $\mathbf{H}^{\dagger}=\mathbf{V}\left(\begin{array}{cc}
         \mathbf{S}^{-1}&0
    \end{array}\right)\mathbf{U}^{\top}$.

    Denote $\mathbf{\bar{H}}=\mathbf{L}\mathbf{H}$, then we have 
    \begin{equation}
        ||\mathbf{\bar{H}}-\mathbf{H}\mathbf{W}||_F^2=\sum_{i=1}^d ||\mathbf{\bar{H}}_{:,i}-\mathbf{H}\mathbf{W}_{:,i}||_2^2
    \end{equation}
    Let $\mathbf{W}^*=\mathbf{H}^{\dagger}\mathbf{\bar{H}}$ , we have $\mathbf{W}^*_{:,i}=\mathbf{H}^{\dagger}\mathbf{\bar{H}}_{:,i}$. 
    Then
    \begin{equation}
    \begin{aligned}
    \mathbf{H}\mathbf{W}^*_{:,i}=&\mathbf{H}\mathbf{H}^{\dagger}\mathbf{\bar{H}}_{:,i}\\
    =&\mathbf{U}\left( \begin{array}{cc}
        I_d & 0 \\
        0 & 0_{n-d}
    \end{array}  \right)\mathbf{U}^{\top}  \mathbf{\bar{H}}_{:,i}
    \end{aligned}
    \end{equation}
    Then
    \begin{equation}
    \begin{aligned}
        ||\mathbf{\bar{H}}_{:,i}-\mathbf{H}\mathbf{W}^*_{:,i}||_2=&||\mathbf{\bar{H}}_{:,i}-\mathbf{H}\mathbf{H}^{\dagger}\mathbf{\bar{H}}_{:,i}||_2\\
        \leq&||I_{n}-\mathbf{H}\mathbf{H}^{\dagger}||_2||\mathbf{\bar{H}}_{:,i}||_2\\
        \leq&||\mathbf{\bar{H}}_{:,i}||_2\\
        =& ||\mathbf{L}\mathbf{H}_{:,i} ||_2\\
        \leq& ||\mathbf{L}||_2 ||\mathbf{H}_{:,i} ||_2\\
        =& \lambda_{\max} (\mathbf{L}) ||\mathbf{H}_{:,i} ||_2
    \end{aligned}
    \end{equation}
    Thus we have 
    \begin{equation}
        \begin{aligned}
            ||\mathbf{\bar{H}}-\mathbf{H}\mathbf{W}||_F^2=&\sum_{i=1}^d ||\mathbf{\bar{H}}_{:,i}-\mathbf{H}\mathbf{W}_{:,i}||_2^2\\
            \leq&  \lambda^2_{\max}(\mathbf{L})\sum_{i=1}^d  ||\mathbf{H}_{:,i} ||_2^2\\
            = & \lambda^2_{\max}(\mathbf{L})||\mathbf{H}||_F^2.
        \end{aligned}
    \end{equation}
    Finally we get
    \begin{equation}
        \begin{aligned}
            \frac{||\mathbf{L}\mathbf{H}-\mathbf{H}\mathbf{W}^*||_F}{||\mathbf{H}||_F}  \leq  \lambda_{\max}(\mathbf{L})
        \end{aligned}
    \end{equation}
    
\end{proof}

\subsection{Implementing Other GNNs as Teacher}\label{app:otherTeacherGNNs}

Here we explain how to implement other GNNs as teacher in \algo, especially on how to apply teacher injection in The Proposed Model section to different GNNs.

\textbf{\gcn.} The message passing of \gcn in the $l$-th layer is 
\begin{equation}\label{eq:gcn}
\begin{aligned}
\gp^\la\text{: } &\tilde{\h}_{v}^\la = \hat{\LM}\HM^{(l-1)},\\
    \ft^\la\text{: } &\h_v^\la = \sigma \left(\tilde{\h}_{v}^\la\cdot \WM^\la +\mathbf{b}^\la \right).
\end{aligned}
\end{equation}
where $\hat{\LM}=\hat{\DM}^{-\frac{1}{2}}\hat{\adj}\hat{\DM}^{-\frac{1}{2}}$ is the normalized Laplacian of the graph, $\hat{\adj}=\adj+I_\numV$ is the adjacency matrix with self loops, $\hat{\DM}_{ii}=\sum_{j}\hat{\adj}_{ij}$ is the degree matrix, $\WM^\la$ is the transformation matrix of $l$-th layer, and $\mathbf{H}^\la$ is the output of $l$-th layer.

For the $l$-th layer of the \gcn teacher, we directly inject $\WM^\la$ and $\mathbf{b}^\la$ of the teacher in Equation \eqref{eq:gcn} into the student FC layer $\fc^\latwo$ in Equation \eqref{eq:studentlayers}, i.e., $\WM_M^\latwo=\WM^\la$ and $\mathbf{b}_M^\latwo=\mathbf{b}^\la$, while using $\fc^\laone$ in  Equation \eqref{eq:studentlayers} to simulate  $\gp^\la$ in Equation \eqref{eq:gcn}.

\textbf{\gat.} The message passing of \gat in the $l$-th layer with single head attention is  
\begin{equation}\label{eq:gat}
\begin{aligned}
\gp^\la\text{: } &\tilde{\h}_{v}^\la =\Pi^\la\HM^{(l-1)},\\
    \ft^\la\text{: } &\h_v^\la = \sigma \left(\tilde{\h}_{v}^\la\cdot \WM^\la +\mathbf{b}^\la \right).
\end{aligned}
\end{equation}
where $\Pi^\la$ is the attention matrix defined as $\Pi_{ij}^\la=\text{LeakyReLU}([\HM_i^{(l-1)}\WM^\la||\mathbf{H}_j^{(l-1)}\WM^\la]^\top \mathbf{a}^\la)$ if $(i,j)$ is an edge in $\G$, i.e., $(i,j)\in \mathcal{E}$, and $\Pi_{ij}^\la=0$ otherwise, $||$ denotes the concatenation operation, and $\mathbf{a}^\la\in \mathbb{R}^{2d_l}$ is a learnable attention vector.

For the $l$-th layer of the \gat teacher, we directly inject $\WM^\la$ and $\mathbf{b}^\la$ of the teacher in Equation \eqref{eq:gat} into the student,  i.e., $\WM_M^\latwo=\WM^\la$ and $\mathbf{b}_M^\latwo=\mathbf{b}^\la$, while using $\fc^\laone$ in  Equation \eqref{eq:studentlayers} to simulate  $\gp^\la$ in Equation \eqref{eq:gat}.

\begin{figure*}[!t]
	\centering
		\includegraphics[width=0.9\textwidth]{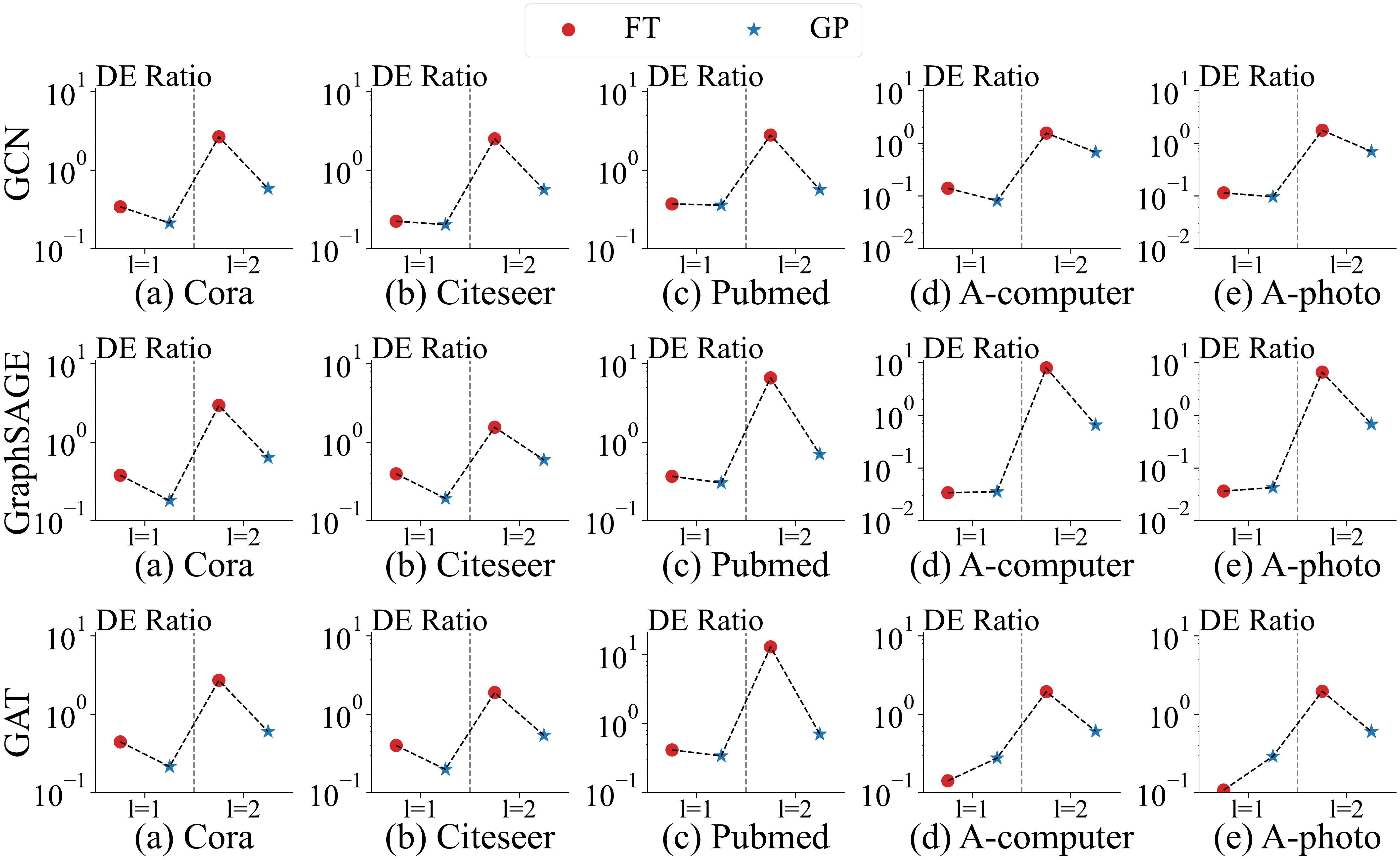}
    \caption{The layer-wise DE ratio of {trained} \gcn, \sage and \gat, with 2 layers architecture on different datasets. The $x$-axis is the layer number, and the $y$-axis is the DE ratio value.}\label{fig:2_layerwise_DE_ratio}
\end{figure*}

\textbf{APPNP.} APPNP decouples feature transformation and graph propagation into two stages.
APPNP first performs FTs  for $T_1$ times in Equation \eqref{eq:appnpTrans} to get $\HM^{(T_1)}$ (usually $T_1=2$), and then propagates $\HM^{(T_1)}$ over the graph for $T_2$ hops in Equation \eqref{eq:appnpProp} by approximate Personalized PageRank.
\begin{equation}\label{eq:appnpTrans}
   \ft^\la\text{: }\HM^\la=  \HM^{(l-1)}\WM^\la, l=1,2,...,T_1, 
\end{equation}
\begin{equation}\label{eq:appnpProp}
   \gp\text{: } \mathbf{H}^\la= (1-\alpha)\hat{\mathbf{L}}\mathbf{H}^\la+\alpha\mathbf{H}^{(T_1)}, l=T_1+1,T_1+2,...,T_1+T_2,
\end{equation}
where $\alpha$ is a hyperparameter and $\hat{\mathbf{L}}$ is the normalized Laplacian mentioned before.

For each of $\ft^\la$ in APPNP, we inject it into a $\fc^l$ layer in the student MLP. In other words, we have $T_1$ FC layers to mimic the FT layers in APPNP.
Then for the $T_2$ steps of propagation in Equation \eqref{eq:appnpProp}, we use one FC layer $\fc^{T_1+1}$ to mimic all of them, since $T_2$ can be large.
In total, there are $T_1+1$ FC layers in MLP to approximate the teacher GNN.
The DED loss of APPNP is 
$\mathcal{L}_{DED}=\sum_{l=1}^{T_1}(\der_{\ft^\la}-\der_{\fc^l})^2+(\der_{\gp}-\der_{\fc^{T_1+1}})^2$

\begin{figure*}[!t]
	\centering
		\includegraphics[width=0.9\textwidth]{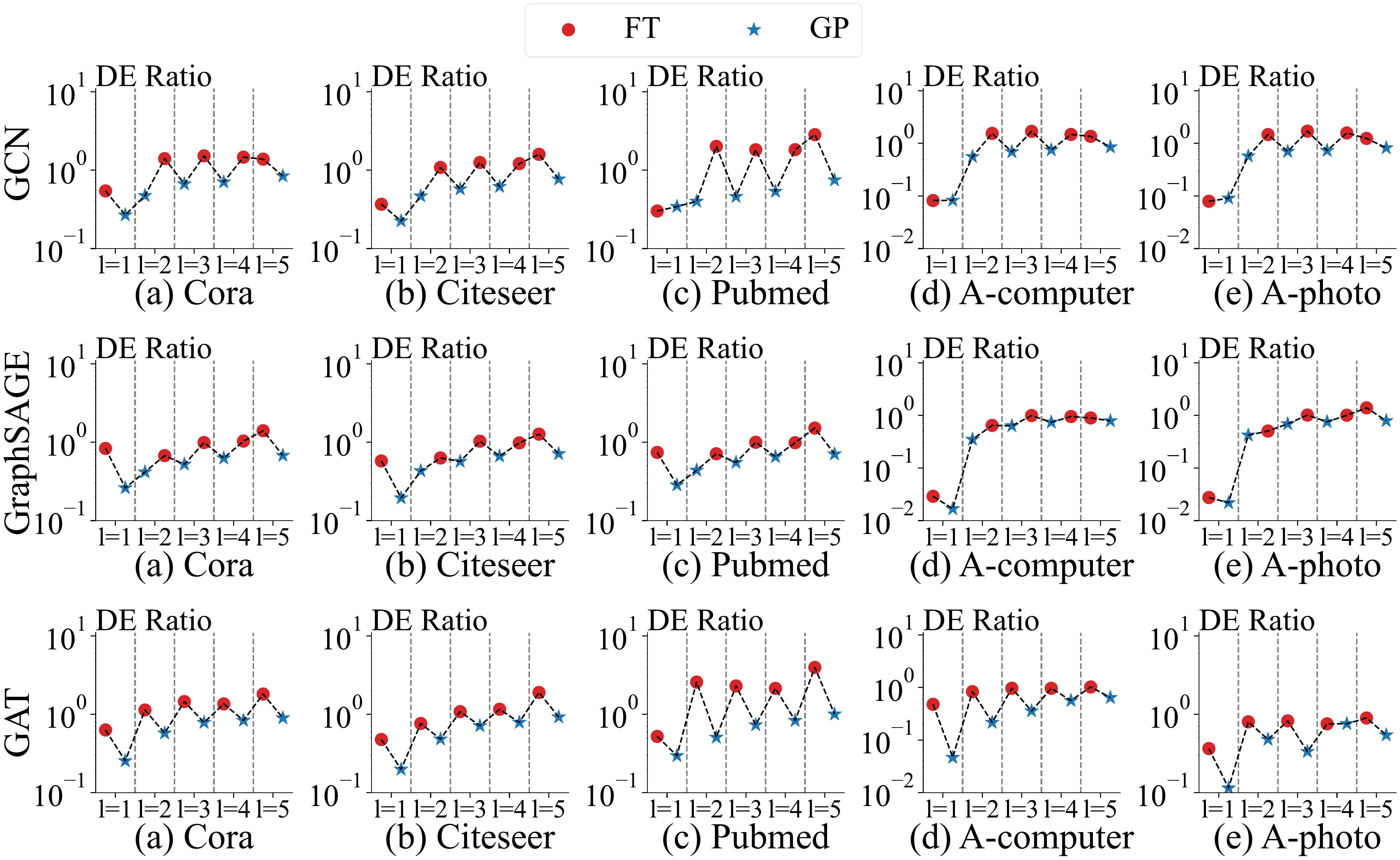}
   \caption{The layer-wise DE ratio of {trained} \gcn, \sage and \gat, with 5 layers architecture on different datasets. The $x$-axis is the layer number, and the $y$-axis is the DE ratio value. }\label{fig:5_layerwise_DE_ratio}
\end{figure*}

\subsection{DE Ratios of other GNN architectures}\label{appendix:deratio}

To show the pattern of DE ratios in different GNN teachers, we plot the layer-wise  DE ratios of \gcn, \sage and \gat, in addition to \sage, with 2 layers architecture on the five benchmarks in Figure~\ref{fig:2_layerwise_DE_ratio}.
From Figure~\ref{fig:2_layerwise_DE_ratio} we can observe that all  GNNs share very similar patterns across datasets: 
within a specific layer, in Figure \ref{fig:2_layerwise_DE_ratio}, we observe the following nearly consistent patterns about \textit{DE-ratio}:
(i) within the same layer for $l=1,2$, most DE ratio $\der_{\ft^\la}$ for $\ft^\la$ is larger than $\der_{\gp^\la}$ of $\gp^\la$, suggesting that $\gp^\la$ actively smooths embeddings, whereas $\ft^\la$ is relatively conservative for smoothing; 
(ii) at $l=2$, DE ratio $\der_{\ft^\la}$ even surpasses 1, indicating that in this layer, $\ft^\la$ acts to diversify embeddings rather than smoothing them.
Furthermore, to investigate the smoothing behavior on deeper layer cases, we plot Figure~\ref{fig:5_layerwise_DE_ratio} showing layer-wise DE ratios of 5-layer  trained GNNs, and 
from Figure~\ref{fig:5_layerwise_DE_ratio} we can observe that
the trained 5-layer GNNs have similar patterns to the 2-layer GNNs: the DE ratios of $\ft^\la$ is generally higher than that of $\gp^\la$, indicating that $\gp^\la$ is aggressive while $\ft^\la$ is relatively conservative for smoothing.

\begin{figure*}[!t]
    \centering
    \includegraphics[width=0.88\linewidth]{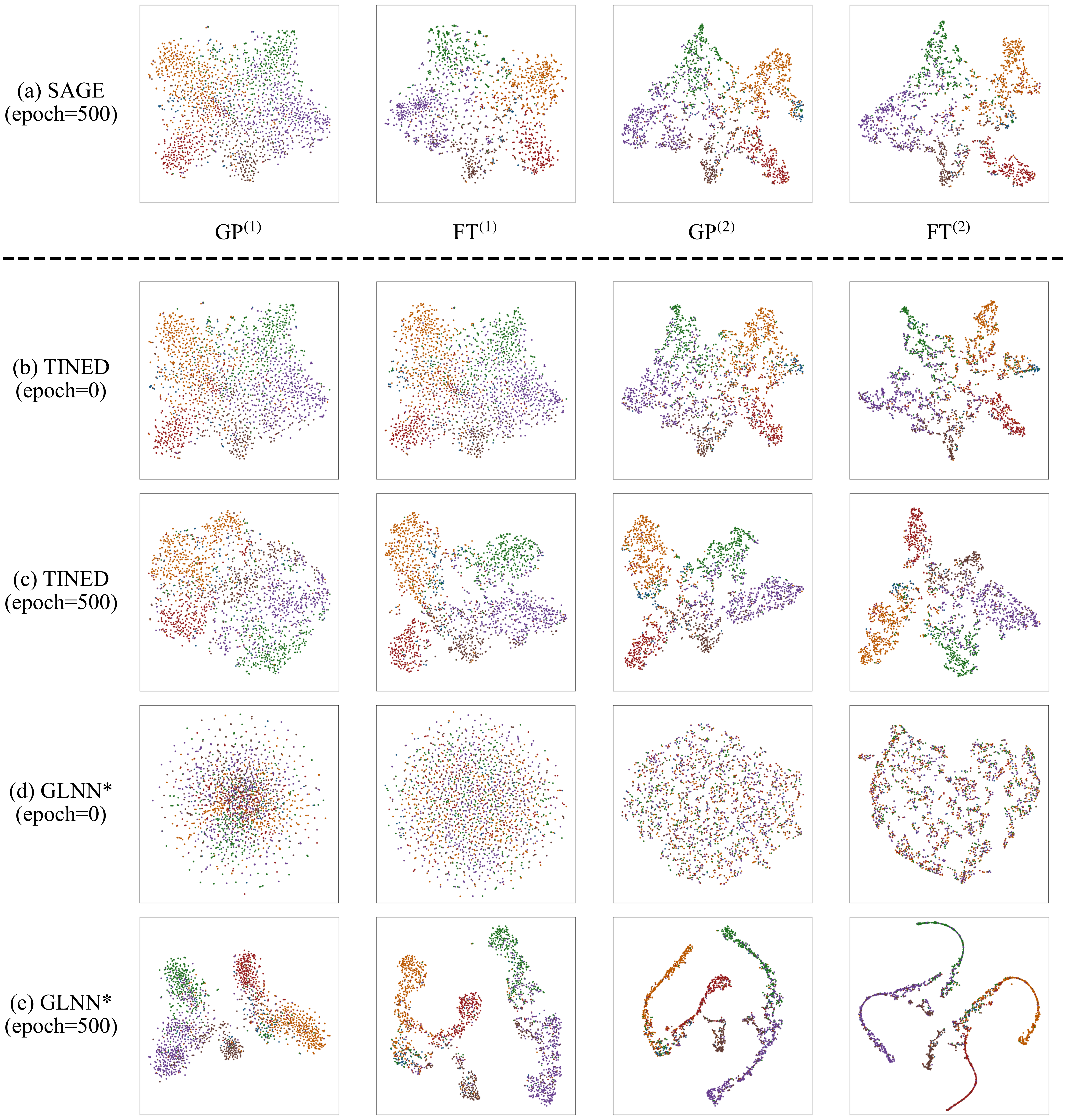}
    \caption{Layer-wise t-SNE visualization of the embeddings for various models (\sage, \algo, \glnnstar) and stages (epoch 0 at initialization and epoch 500 after training convergence) on Citeseer.
  }
    \label{fig:tsne_all}
\end{figure*}

\subsection{Layer-wise t-SNE Visualization}\label{sec:tsne_all}
In Figure~\ref{fig:tsne_all}, we present a detailed layer-wise t-SNE visualization of the embeddings for various models and stages. Figure~\ref{fig:tsne_all}(a) (1st row) shows the 2-layer \sage teacher model. Figure~\ref{fig:tsne_all}(b) (2nd row) illustrates \algo at initialization before training (epoch 0), while Figure~\ref{fig:tsne_all}(c) (3rd row) displays \algo after training convergence (epoch 500). Figure~\ref{fig:tsne_all}(d) (4th row) and Figure~\ref{fig:tsne_all}(e) (5th row) depict the baseline \glnnstar at initialization (epoch 0) and after training convergence (epoch 500), respectively.
For the 2-layer \sage in Figure~\ref{fig:tsne_all}(a), the 1st row visualizes the embeddings from left to right: after the $\ft^{(1)}$ operation in the first layer, the $\gp^{(1)}$ operation in the first layer, $\ft^{(2)}$ in the second layer, and $\gp^{(2)}$ in the second layer. For each distillation method in Figures~\ref{fig:tsne_all}(b-e), each row contains four plots representing the embeddings of the four fully-connected (FC) layers in the methods, corresponding to the four operations in the first row of the teacher.
Compared to the teacher model in Figure~\ref{fig:tsne_all}(a), \algo at initialization in Figure~\ref{fig:tsne_all}(b) already exhibits quite similar embedding patterns, thanks to the proposed teacher injection technique. In contrast, the baseline \glnnstar shows a random visual pattern at epoch 0 in Figure~\ref{fig:tsne_all}(d), which is radically different from the teacher model.
After training convergence at 500 epochs, \algo's visualized embeddings in Figure~\ref{fig:tsne_all}(c) closely resemble those of the teacher in Figure~\ref{fig:tsne_all}(a), whereas the baseline \glnnstar in Figure~\ref{fig:tsne_all}(e) produces embeddings that are visually distinct. These observations demonstrate the effectiveness of the teacher injection and Dirichlet energy distillation techniques in \algo.

\eat{Here we provide a more detailed layer-wised TSNE visualization of Figure~\ref{fig:tsne_part} in Figure~\ref{fig:tsne_all}. In alignment with the observations in previous sections, from Figure~\ref{fig:tsne_all} we can furthermore see that the consistency of \algo and teacher embeddings are still maintained in different layers, while \glnnstar diverges more as layers increase. 
For example, in layer $1$, i.e., column 1, we can see the untrained \glnnstar (epoch=0) has totally random embeddings, while untrained \algo has the same informative representations as teacher \sage.
As the layer deepens, trained \glnnstar shows more and more discrepancy between its teacher, while trained \algo maintains the layer-wise structure of teacher outputs. }

\subsection{Computer Resource Details}\label{app:computer}

The experiments on both baselines and our approach are implemented using PyTorch, the DGL~\citep{wang2019dgl} library for GNN algorithms, and Adam~\citep{adam} for optimization. We run
all experiments on  Intel(R) Xeon(R) Platinum 8338C CPU @ 2.60GHz CPU, and a Nvidia Geforce 3090 Cards with Cuda version
11.7. Source codes of all competitors are obtained from
respective authors. The totally training time cost for one set of hyper-parameters varies among datasets, from \cora costing 10 minutes to \ogbp costing 13 hours.

\begin{table}[t]
\center
\caption{Hyperparameters for GNNs on five datasets from \cite{glnn,nosmog}.}
\small
\begin{tabular}{@{}llll}
\toprule
{Dataset} & \sage& \gcn& \gat\\ \midrule
\# layers& 2& 2& 2\\
hidden dim& 128& 64& 64\\ 
learning rate&  0.01&  0.01&  0.01\\ 
weight decay& 0.0005& 0.001& 0.01\\
dropout&  0& 0.8& 0.6\\
fan out& 5,5& -& -\\
attention heads& -& -& 8\\ 
\bottomrule
\end{tabular}
\label{tab:teacher_hyper}
\end{table}

\subsection{Teacher details and hyperparameter search space} \label{app:teacherhyper}

The hyperparameters of GNN models on each dataset are taken from the best hyperparameters provided by previous studies~\citep{nosmog,glnn}. 
For APPNP teacher model, we found that the suggested hyper-parameter in previous literature~\cite{glnn,nosmog} produces relatively poor results, thus we search it in following space, achieving a better APPNP teacher: \# layers from $[2,3]$, learning rate from $[0.0001,0.001,0.01]$, weight decay from $[0.0001,0.001,0.01]$, dropout from $[0,0.5,0.8]$, hidden dim from $[128,256]$, power iteration $K$ from $[5,10,15]$.
For the students MLP, GLNN and NOSMOG, we set the number of layers and the hidden dimension of each layer to be the same as the teacher GNN, so their total number of parameters stays the same as the teacher GNN. 
For \algo, \algog, \glnnstar and \nosmogstar, given a layer-wise GNN teacher with $T$ layers (including, \sage, \gcn, and \gat), in the student MLP we set the same hidden dimension as teachers, and $2T$ layers, while for APPNP,  we set the student MLP to have $T_1+1$ layers as explained in Appendix \ref{app:otherTeacherGNNs}.
Other hyper-parameter searching spaces are listed here: Learning rate from $[0.0001, 0.0005, 0.001, 0.005, 0.01]$, weight decay from $[0.0, 0.0001, 0.0005, 0.001, 0.005, 0.01]$, weight of distillation $\lambda$ from $[0.1,0.4,0.5,0.6,1]$, nornamlization type from $[\text{batch normalization}, \text{layer normalization}, \text{none}]$, dropout from $[0,0.1,0.3,0.5,0.8]$. Batch size for two large OGB datasets from $[512,1024,4096]$. 
 Weight of DED $\beta$ from $[1e^{-6},5e^{-5},1e^{-5},0.05, 0.1, 0.5, 1, 5, 10 ]$. Fine tuning weight $\eta$ for injected teacher FT layers from $[0.01, 0.1, 0.5, 1,3 ,10]$, 
For the hyperparamrter space of \algog from \nosmog,  the search space is the same as \cite{nosmog}.
In large OGB datasets, the direct computation of DE values will run out of GPU memory 24GB. Thus, we propose a sampling ratio $\zeta$ to compute approximate DE values in DED process. This is done by inducing a subgraph based on random sampling of edges, then approximate the DE ratio of input and output feature matrix on this sampled subgraph. We also search this sampling proportion $\zeta$ from $[0.001, 0.005, 0.1, 0.4, 1]$.
Moreover, empirically we found that applying smoothing term on the DE ratio in MSE loss could benefit optimization, thus we propose a smoothing function technique $\mu$ on DE ratio $\der_{op}$, i.e. $\mu{(\der_{op}})$, when computing the DED loss, where $\mu$ is searched from $[\text{sqrt}(\cdot),\text{log}(\cdot),\text{Identity}(\cdot)]$~\citep{msle_survey}.

\subsection{Impact of $\zeta$ on Large Datasets}
The parameter $\zeta$ controls the subgraph sampling ratio used to estimate the DE ratio, helping to avoid memory overflow on large datasets. We vary $\zeta$ and report the results on the large ogbn-arxiv dataset in Table \ref{tab:vary_zeta_bbbm}. The results show that TINED maintains stable performance across different $\zeta$ values. Notably, $\zeta$ values larger than 0.8 are unnecessary, as they increase computational overhead without improving distillation quality.

\begin{table}[t]
  \centering
  \caption{Vary $\zeta$ for \algo on large graph ogbn-arxiv under transductive setting.}
  \label{tab:vary_zeta_bbbm}
  \begin{adjustbox}{width=0.95\linewidth,center}
  \begin{tabular}{@{}llllll@{}}
    \toprule
    $\zeta$ & 0.1 & 0.3 & 0.5 & 0.7 & $>$0.8 \\
    \midrule
    \ogba & 64.31$\pm$0.71 & 64.32$\pm$0.91 & 64.38$\pm$0.75 & {64.44$\pm$0.80} & OOM \\
    \bottomrule
  \end{tabular}
  \end{adjustbox}
    \vspace{-3mm}
\end{table}

\begin{table}[t]
    \centering
    \caption{Results of VQGraph and our \algog with \sage as teacher on transductive setting. 
    The best result is in bold.
    }
    \label{tab:my_label}
  \begin{adjustbox}{width=1\linewidth,center}
    \setlength{\tabcolsep}{3pt}
    \begin{tabular}{cccccc}
    \toprule
         & \cora & \citseer & \pubmed& \acomputer & \aphotos \\
         \midrule
         VQGraph &78.66$\pm$1.21&74.66$\pm$1.23&73.02$\pm$3.51&80.16$\pm$2.02&92.32$\pm$1.74\\
         \algog & \textbf{83.70$\pm$1.02} &\textbf{75.39$\pm$1.59} &\textbf{77.75$\pm$3.14} &\textbf{84.82$\pm$1.58} &\textbf{94.05$\pm$0.39} \\
         \bottomrule
    \end{tabular}
    \end{adjustbox}
    \vspace{-3mm}
\end{table}

\subsection{Comparison with VQGraph}

In this section, we include a comparison with VQGraph~\cite{vqgraph} with \sage teacher.
We thoroughly searched all the hyperparameter spaces specified in the original VQGraph paper (Table 12 of its paper). 
Specifically, VQGraph has the following hyperparameter search space for the teacher \sage with codebook: max epoch $\in$ [100, 200, 500], hidden dim $=$ 128, dropout\ ratio $\in$ [0, 0.2, 0.4, 0.6, 0.8], learning  rate $\in$ [0.01, 1e-3, 1e-4], weight decay $\in$ [1e-3, 5e-4, 0], codebook size $\in$ [8192, 16384], lamb node $\in$ [0, 0.01, 0.001, 1e-4], and lamb edge $\in$ [0, 1e-1, 0.03, 0.01, 1e-3]. 
For distillation, the hyperparameter search space of VQGraph is: max epoch $\in$ [200, 500], norm type $\in$ [“batch”, “layer”, “none”], hidden dim $=$ 128, dropout ratio $\in$ [0, 0.1, 0.4, 0.5, 0.6], learning rate $\in$ [0.01, 5e-3, 3e-3, 1e-3], weight decay $\in$ [5e-3, 1e-3, 1e-4, 0], lamb soft labels $\in$ [0.5, 1], and lamb soft tokens $\in$ [1e-8, 1e-3, 1e-1, 1]. 
The table below reports the results of VQGraph and our method \algog. 
Observe that our method outperforms VQGraph on the datasets, which validates the effectiveness of \algo.

Since VQGraph involves re-training the teacher, whereas all other competitors, including KRD, FF-G2M, GLNN, NOSMOG, and \algo, use a fixed teacher for distillation, we believe VQGraph belongs to a different category and have therefore excluded it from the main result.

%% file: example_paper.bbl
\begin{thebibliography}{51}
\providecommand{\natexlab}[1]{#1}
\providecommand{\url}[1]{\texttt{#1}}
\expandafter\ifx\csname urlstyle\endcsname\relax
  \providecommand{\doi}[1]{doi: #1}\else
  \providecommand{\doi}{doi: \begingroup \urlstyle{rm}\Url}\fi

\bibitem[Chang et~al.(2022)Chang, Yang, and Lee]{layerwise2}
Heng-Jui Chang, Shu-wen Yang, and Hung-yi Lee.
\newblock Distilhubert: Speech representation learning by layer-wise distillation of hidden-unit bert.
\newblock In \emph{ICASSP}, pages 7087--7091. IEEE, 2022.

\bibitem[Chen et~al.(2020{\natexlab{a}})Chen, Lin, Li, Li, Zhou, and Sun]{measuringSmoothingAndRelieving}
Deli Chen, Yankai Lin, Wei Li, Peng Li, Jie Zhou, and Xu~Sun.
\newblock Measuring and relieving the over-smoothing problem for graph neural networks from the topological view.
\newblock In \emph{AAAI}, pages 3438--3445, 2020{\natexlab{a}}.

\bibitem[Chen et~al.(2021)Chen, Chen, and Bruna]{GA_MLP}
Lei Chen, Zhengdao Chen, and Joan Bruna.
\newblock On graph neural networks versus graph-augmented mlps.
\newblock In \emph{ICLR}, 2021.

\bibitem[Chen et~al.(2020{\natexlab{b}})Chen, Wei, Huang, Ding, and Li]{GCNII}
Ming Chen, Zhewei Wei, Zengfeng Huang, Bolin Ding, and Yaliang Li.
\newblock Simple and deep graph convolutional networks.
\newblock In \emph{ICML}, 2020{\natexlab{b}}.

\bibitem[Deng and Zhang(2021)]{GFKD}
Xiang Deng and Zhongfei Zhang.
\newblock Graph-free knowledge distillation for graph neural networks.
\newblock \emph{IJCAI}, 2021.

\bibitem[Ding et~al.(2024)Ding, Shi, Li, and Cao]{DZH2024}
Zhihao Ding, Jieming Shi, Qing Li, and Jiannong Cao.
\newblock Effective illicit account detection on large cryptocurrency multigraphs.
\newblock In \emph{CIKM}, page 457–466, 2024.

\bibitem[Dong et~al.(2024)Dong, Zhang, and Wang]{dong2023rayleigh}
Xiangyu Dong, Xingyi Zhang, and Sibo Wang.
\newblock Rayleigh quotient graph neural networks for graph-level anomaly detection.
\newblock In \emph{ICLR}, 2024.

\bibitem[Dong et~al.(2025{\natexlab{a}})Dong, Zhang, Chen, Yuan, and Wang]{dong2025spacegnn}
Xiangyu Dong, Xingyi Zhang, Lei Chen, Mingxuan Yuan, and Sibo Wang.
\newblock Spacegnn: Multi-space graph neural network for node anomaly detection with extremely limited labels.
\newblock In \emph{ICLR}, 2025{\natexlab{a}}.

\bibitem[Dong et~al.(2025{\natexlab{b}})Dong, Zhang, Sun, Chen, Yuan, and Wang]{dong2025smoothgnn}
Xiangyu Dong, Xingyi Zhang, Yanni Sun, Lei Chen, Mingxuan Yuan, and Sibo Wang.
\newblock Smoothgnn: Smoothing-aware gnn for unsupervised node anomaly detection.
\newblock In \emph{Proceedings of the ACM on Web Conference 2025}, pages 1225--1236, 2025{\natexlab{b}}.

\bibitem[Feng et~al.(2022)Feng, Li, Yuan, and Wang]{freekd}
Kaituo Feng, Changsheng Li, Ye~Yuan, and Guoren Wang.
\newblock Freekd: Free-direction knowledge distillation for graph neural networks.
\newblock In \emph{KDD}, 2022.

\bibitem[Hamilton et~al.(2017)Hamilton, Ying, and Leskovec]{graphsage}
William~L. Hamilton, Rex Ying, and Jure Leskovec.
\newblock Inductive representation learning on large graphs.
\newblock In \emph{NeurIPS}, 2017.

\bibitem[Han et~al.(2023)Han, Zhao, Liu, Hu, and Shah]{mlpinit}
Xiaotian Han, Tong Zhao, Yozen Liu, Xia Hu, and Neil Shah.
\newblock Mlpinit: Embarrassingly simple {GNN} training acceleration with {MLP} initialization.
\newblock In \emph{ICLR}, 2023.

\bibitem[Hinton et~al.(2015)Hinton, Vinyals, and Dean]{kd_hinton}
Geoffrey~E. Hinton, Oriol Vinyals, and Jeffrey Dean.
\newblock Distilling the knowledge in a neural network.
\newblock \emph{arXiv preprint arXiv:2211.02989}, 2015.

\bibitem[Hu et~al.(2020)Hu, Fey, Zitnik, Dong, Ren, Liu, Catasta, and Leskovec]{ogb}
Weihua Hu, Matthias Fey, Marinka Zitnik, Yuxiao Dong, Hongyu Ren, Bowen Liu, Michele Catasta, and Jure Leskovec.
\newblock Open graph benchmark: Datasets for machine learning on graphs.
\newblock In \emph{NeurIPS}, 2020.

\bibitem[Hu et~al.(2021)Hu, You, Wang, Wang, Zhou, and Gao]{graph_mlp}
Yang Hu, Haoxuan You, Zhecan Wang, Zhicheng Wang, Erjin Zhou, and Yue Gao.
\newblock Graph-mlp: node classification without message passing in graph.
\newblock \emph{arXiv preprint arXiv:2106.04051}, 2021.

\bibitem[Jadon et~al.(2022)Jadon, Patil, and Jadon]{msle_survey}
Aryan Jadon, Avinash Patil, and Shruti Jadon.
\newblock A comprehensive survey of regression based loss functions for time series forecasting.
\newblock \emph{arXiv preprint arXiv:2211.02989}, 2022.

\bibitem[Jia et~al.(2020)Jia, Lin, Ying, You, Leskovec, and Aiken]{jia2020redundancy}
Zhihao Jia, Sina Lin, Rex Ying, Jiaxuan You, Jure Leskovec, and Alex Aiken.
\newblock Redundancy-free computation for graph neural networks.
\newblock In \emph{KDD}, 2020.

\bibitem[Jiao et~al.(2020)Jiao, Yin, Shang, Jiang, Chen, Li, Wang, and Liu]{kd_jiao}
Xiaoqi Jiao, Yichun Yin, Lifeng Shang, Xin Jiang, Xiao Chen, Linlin Li, Fang Wang, and Qun Liu.
\newblock Tinybert: Distilling {BERT} for natural language understanding.
\newblock In \emph{EMNLP}, pages 4163--4174. Association for Computational Linguistics, 2020.

\bibitem[Kingma and Ba(2015)]{adam}
Diederik~P. Kingma and Jimmy Ba.
\newblock Adam: {A} method for stochastic optimization.
\newblock In \emph{ICLR}, 2015.

\bibitem[Kipf and Welling(2017)]{gcn}
Thomas~N. Kipf and Max Welling.
\newblock Semi-supervised classification with graph convolutional networks.
\newblock In \emph{ICLR}, 2017.

\bibitem[Klicpera et~al.(2019)Klicpera, Bojchevski, and G{\"u}nnemann]{appnp}
Johannes Klicpera, Aleksandar Bojchevski, and Stephan G{\"u}nnemann.
\newblock Predict then propagate: Graph neural networks meet personalized pagerank.
\newblock In \emph{ICLR}, 2019.

\bibitem[Lee and Song(2019)]{lee2019graph}
Seunghyun Lee and Byung~Cheol Song.
\newblock Graph-based knowledge distillation by multi-head attention network.
\newblock \emph{arXiv preprint arXiv:1907.02226}, 2019.

\bibitem[Li et~al.(2019)Li, Lin, Chen, and Chiang]{layerwise3}
Hao-Ting Li, Shih-Chieh Lin, Cheng-Yeh Chen, and Chen-Kuo Chiang.
\newblock Layer-level knowledge distillation for deep neural network learning.
\newblock \emph{Applied Sciences}, 9\penalty0 (10):\penalty0 1966, 2019.

\bibitem[Li et~al.(2020)Li, Wang, Wang, and Leskovec]{distance_encoding}
Pan Li, Yanbang Wang, Hongwei Wang, and Jure Leskovec.
\newblock Distance encoding: Design provably more powerful neural networks for graph representation learning.
\newblock In \emph{NeurIPS}, 2020.

\bibitem[Liang et~al.(2023)Liang, Zuo, Zhang, He, Chen, and Zhao]{layerwise1}
Chen Liang, Simiao Zuo, Qingru Zhang, Pengcheng He, Weizhu Chen, and Tuo Zhao.
\newblock Less is more: Task-aware layer-wise distillation for language model compression.
\newblock In \emph{ICML}, pages 20852--20867. PMLR, 2023.

\bibitem[Liu et~al.(2020)Liu, Gao, and Ji]{liu2020towards}
Meng Liu, Hongyang Gao, and Shuiwang Ji.
\newblock Towards deeper graph neural networks.
\newblock In \emph{KDD}, pages 338--348, 2020.

\bibitem[Platonov et~al.(2023)Platonov, Kuznedelev, Diskin, Babenko, and Prokhorenkova]{platonov2023critical}
Oleg Platonov, Denis Kuznedelev, Michael Diskin, Artem Babenko, and Liudmila Prokhorenkova.
\newblock A critical look at the evaluation of gnns under heterophily: Are we really making progress?
\newblock \emph{ICLR}, 2023.

\bibitem[Rozemberczki et~al.(2021)Rozemberczki, Allen, and Sarkar]{squirrel}
Benedek Rozemberczki, Carl Allen, and Rik Sarkar.
\newblock Multi-scale attributed node embedding.
\newblock \emph{Journal of Complex Networks}, 9\penalty0 (2):\penalty0 cnab014, 2021.

\bibitem[Rusch et~al.(2023)Rusch, Bronstein, and Mishra]{oversmoothingDEmeasure}
T~Konstantin Rusch, Michael~M Bronstein, and Siddhartha Mishra.
\newblock A survey on oversmoothing in graph neural networks.
\newblock \emph{arXiv preprint arXiv:2303.10993}, 2023.

\bibitem[Tian et~al.(2022)Tian, Zhang, Guo, Zhang, and Chawla]{nosmog}
Yijun Tian, Chuxu Zhang, Zhichun Guo, Xiangliang Zhang, and Nitesh Chawla.
\newblock Learning mlps on graphs: A unified view of effectiveness, robustness, and efficiency.
\newblock In \emph{ICLR}, 2022.

\bibitem[Tian et~al.(2023)Tian, Pei, Zhang, Zhang, and Chawla]{kg_on_graph_survey}
Yijun Tian, Shichao Pei, Xiangliang Zhang, Chuxu Zhang, and Nitesh~V Chawla.
\newblock Knowledge distillation on graphs: A survey.
\newblock \emph{arXiv preprint arXiv:2302.00219}, 2023.

\bibitem[Veličković et~al.(2018)Veličković, Cucurull, Casanova, Romero, Liò, and Bengio]{gat}
Petar Veličković, Guillem Cucurull, Arantxa Casanova, Adriana Romero, Pietro Liò, and Yoshua Bengio.
\newblock Graph attention networks.
\newblock In \emph{ICLR}, 2018.

\bibitem[Wang et~al.(2022)Wang, Yin, Zhang, and Li]{peg}
Haorui Wang, Haoteng Yin, Muhan Zhang, and Pan Li.
\newblock Equivariant and stable positional encoding for more powerful graph neural networks.
\newblock In \emph{ICLR}, 2022.

\bibitem[Wang et~al.(2019)Wang, Zheng, Ye, Gan, Li, Song, Zhou, Ma, Yu, Gai, Xiao, He, Karypis, Li, and Zhang]{wang2019dgl}
Minjie Wang, Da~Zheng, Zihao Ye, Quan Gan, Mufei Li, Xiang Song, Jinjing Zhou, Chao Ma, Lingfan Yu, Yu~Gai, Tianjun Xiao, Tong He, George Karypis, Jinyang Li, and Zheng Zhang.
\newblock Deep graph library: A graph-centric, highly-performant package for graph neural networks.
\newblock \emph{arXiv preprint arXiv:1909.01315}, 2019.

\bibitem[Wu et~al.(2023{\natexlab{a}})Wu, Lin, Huang, Fan, and Li]{ffg2m}
Lirong Wu, Haitao Lin, Yufei Huang, Tianyu Fan, and Stan~Z. Li.
\newblock Extracting low-/high- frequency knowledge from graph neural networks and injecting it into mlps: An effective gnn-to-mlp distillation framework.
\newblock In \emph{AAAI}, pages 10351--10360, 2023{\natexlab{a}}.

\bibitem[Wu et~al.(2023{\natexlab{b}})Wu, Lin, Huang, and Li]{krd}
Lirong Wu, Haitao Lin, Yufei Huang, and Stan~Z. Li.
\newblock Quantifying the knowledge in gnns for reliable distillation into mlps.
\newblock In \emph{ICML}, pages 37571--37581, 2023{\natexlab{b}}.

\bibitem[Wu et~al.(2020)Wu, Pan, Chen, Long, Zhang, and Philip]{gnn_survey_1}
Zonghan Wu, Shirui Pan, Fengwen Chen, Guodong Long, Chengqi Zhang, and S~Yu Philip.
\newblock A comprehensive survey on graph neural networks.
\newblock \emph{IEEE transactions on neural networks and learning systems}, 2020.

\bibitem[Yan et~al.(2020)Yan, Wang, Guo, and Lou]{tinygnn}
Bencheng Yan, Chaokun Wang, Gaoyang Guo, and Yunkai Lou.
\newblock Tinygnn: Learning efficient graph neural networks.
\newblock In \emph{KDD}, 2020.

\bibitem[Yang et~al.(2021)Yang, Liu, and Shi]{cpf_plp_ft}
Cheng Yang, Jiawei Liu, and Chuan Shi.
\newblock Extract the knowledge of graph neural networks and go beyond it: An effective knowledge distillation framework.
\newblock In \emph{WWW}, 2021.

\bibitem[Yang et~al.(2024)Yang, Tian, Xu, Liu, Hong, Qu, Zhang, CUI, Zhang, and Leskovec]{vqgraph}
Ling Yang, Ye~Tian, Minkai Xu, Zhongyi Liu, Shenda Hong, Wei Qu, Wentao Zhang, Bin CUI, Muhan Zhang, and Jure Leskovec.
\newblock Vqgraph: Rethinking graph representation space for bridging gnns and mlps.
\newblock In \emph{ICLR}, 2024.

\bibitem[Yang et~al.(2020)Yang, Qiu, Song, Tao, and Wang]{LSP}
Yiding Yang, Jiayan Qiu, Mingli Song, Dacheng Tao, and Xinchao Wang.
\newblock Distilling knowledge from graph convolutional networks.
\newblock In \emph{CVPR}, 2020.

\bibitem[You et~al.(2019)You, Ying, and Leskovec]{pgnn}
Jiaxuan You, Rex Ying, and Jure Leskovec.
\newblock Position-aware graph neural networks.
\newblock In \emph{ICML}, 2019.

\bibitem[Zhang et~al.(2022{\natexlab{a}})Zhang, Huang, Tian, Wen, Ouyang, Li, Ye, and Zhang]{graphdec}
Chunhui Zhang, Chao Huang, Yijun Tian, Qianlong Wen, Zhongyu Ouyang, Youhuan Li, Yanfang Ye, and Chuxu Zhang.
\newblock Diving into unified data-model sparsity for class-imbalanced graph representation learning.
\newblock In \emph{GLFrontiers}, 2022{\natexlab{a}}.

\bibitem[Zhang et~al.(2019)Zhang, Song, Huang, Swami, and Chawla]{hetgnn}
Chuxu Zhang, Dongjin Song, Chao Huang, Ananthram Swami, and Nitesh~V Chawla.
\newblock Heterogeneous graph neural network.
\newblock In \emph{KDD}, pages 793--803, 2019.

\bibitem[Zhang et~al.(2020)Zhang, Huang, Liu, Hu, Song, Ge, Zhang, Wang, Zhou, Shuang, et~al.]{agl}
Dalong Zhang, Xin Huang, Ziqi Liu, Zhiyang Hu, Xianzheng Song, Zhibang Ge, Zhiqiang Zhang, Lin Wang, Jun Zhou, Yang Shuang, et~al.
\newblock Agl: A scalable system for industrial-purpose graph machine learning.
\newblock \emph{arXiv preprint arXiv:2003.02454}, 2020.

\bibitem[Zhang et~al.(2022{\natexlab{b}})Zhang, Liu, Sun, and Shah]{glnn}
Shichang Zhang, Yozen Liu, Yizhou Sun, and Neil Shah.
\newblock Graph-less neural networks: Teaching old mlps new tricks via distillation.
\newblock In \emph{ICLR}, 2022{\natexlab{b}}.

\bibitem[Zheng et~al.(2022)Zheng, Huang, Rao, Katariya, Wang, and Subbian]{cold_brew}
Wenqing Zheng, Edward~W Huang, Nikhil Rao, Sumeet Katariya, Zhangyang Wang, and Karthik Subbian.
\newblock Cold brew: Distilling graph node representations with incomplete or missing neighborhoods.
\newblock In \emph{ICLR}, 2022.

\bibitem[Zhou et~al.(2020)Zhou, Cui, Hu, Zhang, Yang, Liu, Wang, Li, and Sun]{gnn_survey_2}
Jie Zhou, Ganqu Cui, Shengding Hu, Zhengyan Zhang, Cheng Yang, Zhiyuan Liu, Lifeng Wang, Changcheng Li, and Maosong Sun.
\newblock Graph neural networks: A review of methods and applications.
\newblock \emph{AI Open}, 2020.

\bibitem[Zhou et~al.(2023{\natexlab{a}})Zhou, Shi, Yang, Zou, and Li]{ZhouSYZ023}
Ziang Zhou, Jieming Shi, Renchi Yang, Yuanhang Zou, and Qing Li.
\newblock Slotgat: Slot-based message passing for heterogeneous graphs.
\newblock In \emph{{ICML}}, volume 202, pages 42644--42657, 2023{\natexlab{a}}.

\bibitem[Zhou et~al.(2023{\natexlab{b}})Zhou, Shi, Zhang, Huang, and Li]{ZhouSZH023}
Ziang Zhou, Jieming Shi, Shengzhong Zhang, Zengfeng Huang, and Qing Li.
\newblock Effective stabilized self-training on few-labeled graph data.
\newblock \emph{Inf. Sci.}, 631:\penalty0 369--384, 2023{\natexlab{b}}.

\bibitem[Zhu et~al.(2021)Zhu, Wang, Shi, Ji, and Cui]{unifiedframework}
Meiqi Zhu, Xiao Wang, Chuan Shi, Houye Ji, and Peng Cui.
\newblock Interpreting and unifying graph neural networks with an optimization framework.
\newblock In \emph{WWW}, page 1215–1226, 2021.

\end{thebibliography}
